%% file: main.tex
\PassOptionsToPackage{table,dvipsnames}{xcolor}	
\documentclass[11pt]{article}
\pdfoutput=1
\usepackage[final]{acl}
\usepackage[dvipsnames]{xcolor}
\usepackage{colortbl}
\usepackage{pdflscape}
\usepackage{tabularx}
\usepackage{longtable}

\usepackage{latexsym}
\usepackage{todonotes}
\usepackage[T1, T5]{fontenc}
\usepackage{CJKutf8}
\usepackage{times}

\AtBeginDocument{\fontencoding{T1}\selectfont}

\newcommand{\vi}[1]{{\fontencoding{T5}\selectfont #1}}

\usepackage[utf8]{inputenc}
\usepackage{hyperref}

\usepackage{microtype}

\usepackage{inconsolata}
\usepackage{booktabs}

\usepackage[listings, skins]{tcolorbox}
\usepackage{listings}
\usepackage{array}
\usepackage{fontawesome5}
\usepackage{graphicx}
\usepackage{multirow} 
\usepackage{makecell}
\usepackage{xspace}
\usepackage{amsmath}
\usepackage{amssymb}
\newcommand{\eqcont}{\textsuperscript{\scalebox{1.0}{$\bigstar$}}}

\lstdefinelanguage{Markdown}{
  basicstyle=\ttfamily\footnotesize,
  sensitive=false,
  morecomment=[l]{\#},   
  morecomment=[s]{```}{```},
  morestring=[b]",        
  morestring=[b]', 
}

\newcommand{\dataname}{\textsc{Owl}\xspace}

\author{}
\title{\raisebox{-0.3em}{\includegraphics[height=2em]{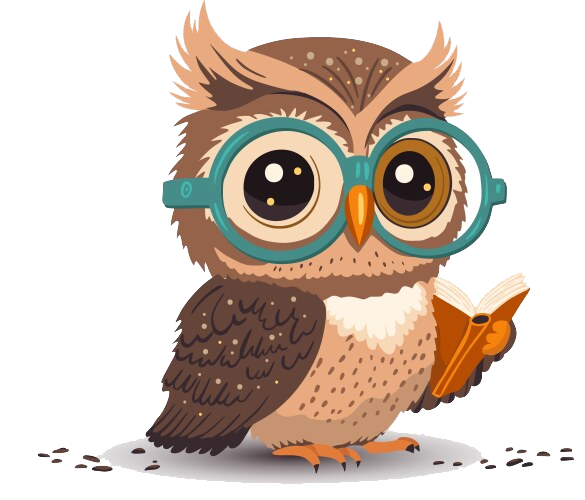}}%
  \hspace{0.3em} OWL: Probing Cross-Lingual Recall of Memorized Texts \\ via World Literature}

\author{
  Alisha Srivastava\eqcont\textsuperscript{\faMagic}\quad
  Emir  Kaan Korukluoglu\eqcont\textsuperscript{\faMagic}\quad
  Minh Nhat Le\eqcont\textsuperscript{\faMagic}\quad
  Duyen Tran\textsuperscript{\faMagic}\quad \\
  \textbf{Chau Minh Pham}\textsuperscript{\faDragon}\quad
  \textbf{Marzena Karpinska}\textsuperscript{\faMoon}\quad 
  \textbf{Mohit Iyyer}\textsuperscript{\faDragon} \\
  \textsuperscript{\faMagic}University of Massachusetts Amherst \quad
  \textsuperscript{\faDragon}University of Maryland, College Park \\ 
  \textsuperscript{\faMoon}Microsoft \\
  \texttt{\{alishasrivas, ekorukluoglu, nhatminhle, duyent\}@umass.edu}\\
\texttt{\{chau\textsuperscript{\faEnvelope}, miyyer\}@umd.edu}, \texttt{mkarpinska@microsoft.com}\textsuperscript{\faEnvelope}
}

\begin{document}
\tracingfonts=1
\maketitle

\def\thefootnote{\eqcont}\footnotetext{These authors contributed equally to this work.}\def\thefootnote{\arabic{footnote}}
\def\thefootnote{\faEnvelope}\footnotetext{Corresponding authors}\def\thefootnote{\arabic{footnote}}

\input{sections/0-abstract}
\input{sections/1-intro}
\input{sections/2-data-collection}
\input{sections/3-experiments}
\input{sections/4-analysis}
\input{sections/5-related-work}
\input{sections/6-discussions_conclusion}
\input{sections/7-limitations}
\input{sections/8-ethics}

\bibliography{references}
\newpage  
\appendix
\input{sections/9-appendix}
\end{document}

%% file: sections/0-abstract.tex
\begin{abstract}

Large language models (LLMs) are known to memorize and recall English text from their pretraining data. However, the extent to which this ability generalizes to other languages or transfers across languages remains unclear.
This paper investigates multilingual and
cross-lingual memorization in LLMs, probing whether memorized content in one language (e.g., English) can be recalled when presented in a different language. To do so, we introduce \dataname, a dataset of \textbf{31.5K} aligned excerpts from 20 books in ten languages, including original English texts, official translations (Vietnamese, Spanish, Turkish), and new translations in six low-resource languages (Sesotho, Yoruba, Maithili, Malagasy, Setswana, Tahitian). We evaluate memorization across model families and sizes through three tasks: (1) \textit{direct probing}, which asks the model to identify a book's title and author; (2) \textit{name cloze}, which requires predicting masked character names; and (3) \textit{prefix probing}, which involves generating continuations. 
We find that some LLMs consistently recall content across languages, even for texts without existing translation. GPT-4o, for example, identifies authors and titles 69.4\% of the time and masked entities 6.3\% of the time in newly translated excerpts. While perturbations (e.g., masking characters, shuffling words) reduce accuracy, the model's performance remains above chance level. Our results highlight the extent of cross-lingual memorization and provide insights on the differences between the models.

\end{abstract}

\begin{center}
  \faGithub \quad \texttt{\small\href{https://github.com/emirkaan5/OWL}{https://github.com/emirkaan5/OWL}}
\end{center}

%% file: sections/1-intro.tex
\section{Introduction}
\label{sec:introduction}

\begin{figure*}[t]
  \centering
  \includegraphics[width=\textwidth]{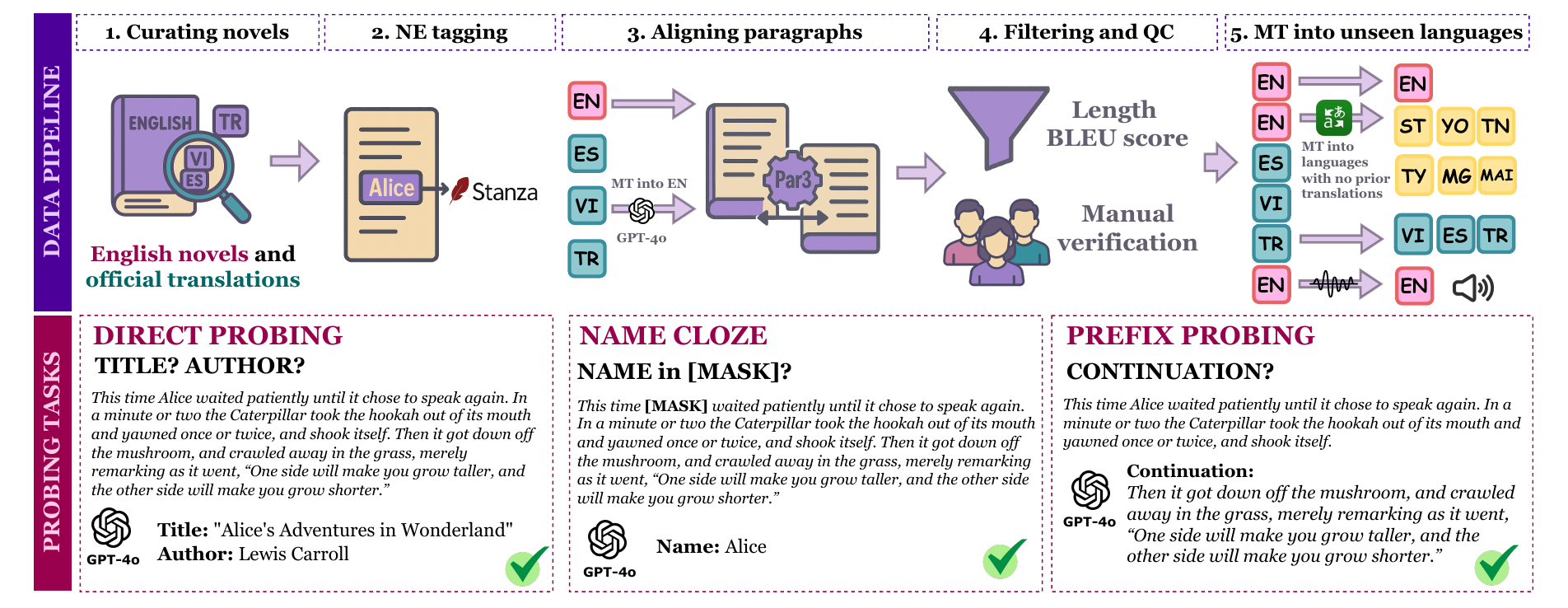}
  \caption{\textbf{Top:} \dataname collection pipeline: (1) Identify English novels with official Turkish, Spanish, and Vietnamese translations; (2) Tag passages with named characters; (3) Align translations to English originals using Par3 aligner \cite{thai2022exploringdocumentlevelliterarymachine}; (4) Filter alignments based on length and BLEU scores, followed by manual verification; (5) Translate validated English passages into six new languages without official translations. \textbf{Bottom:} Probing tasks: (1) Direct Probing (DP) -- identify author/title from a passage; (2) Name Cloze (NC) -- predict masked names in passages; (3) Prefix Probing (PP) -- generate continuations from passage prefixes. Prompt texts omitted for clarity (see \autoref{fig:direct-probing}, \autoref{fig:nct}, \autoref{fig:pp}). The figure shows outputs from GPT-4o. See \autoref{tab:data_overview} for an overview of our experiments.}
  \label{fig:pipeline}
\end{figure*}

Large language models (LLMs) encode substantial factual and linguistic knowledge from their training corpora, which they can later access to respond to user queries \cite{petroni-etal-2019-language, kassner-etal-2021-multilingual}.
Prior work investigating how LLMs acquire and recall this information has primarily focused on English texts~\citep{carlini2021extractingtrainingdatalarge, 9833649, Golchin2023TimeTI, huang2024demystifyingverbatimmemorizationlarge,  shi_detecting_2023, ravichander2025halogenfantasticllmhallucinations}.
Hence, it remains unclear how much content LLMs memorize in languages other than English, and whether such knowledge can be reliably accessed in a language different from the one in which it was originally learned.
While \citet{goldman2025eclekticnovelchallengeset} investigate cross-lingual knowledge transfer, their methodology assumes that content is unseen in a target language if its Wikipedia article is missing.  This assumption is potentially problematic, as the same information may exist in other online sources within the pretraining data.

To address these limitations and investigate multilingual memorization and cross-lingual knowledge recall, we introduce \dataname{}, a new dataset comprising \textbf{31,540} aligned literary passages from \textbf{20} English books. \dataname{} is unique in that it includes not only official human translations in Spanish, Turkish, and Vietnamese, but also \textit{newly produced} machine translations into six low-resource languages (Sesotho, Yoruba, Maithili, Malagasy, Setswana, and Tahitian) for which no published translations of these works previously existed.

Leveraging \dataname{}, we extend the probing methodology of prior work and employ three probing tasks:
(1) \textbf{direct probing}~\cite{karamolegkou-etal-2023-copyright}, where the LLM identifies a book's title and author from a passage; (2) \textbf{name cloze task}~\citep{chang2023speakmemoryarchaeologybooks}, where it fills in a masked character name; and (3) \textbf{prefix probing}~\citep{karamolegkou-etal-2023-copyright, carlini2023quantifying}, where it continues a given passage. These probing tasks allow us to investigate three research questions:

\noindent\textbf{First, we examine the memorization of official translations.} By comparing LLM performance on original English texts (e.g., \emph{Alice in Wonderland}) against their published human translations, we find that while memorization is present across languages, it is more prominent in English. For instance, in direct probing LLMs achieve 63.8\% averaged accuracy for English excerpts versus 47.2\% for Spanish, Turkish, and Vietnamese examples. This multilingual memorization persists even when contextual coherence is disrupted by shuffling words in the passage.

\noindent\textbf{Second, we quantify cross-lingual memorization using our newly produced translations.} Since these translations are novel and the original works lack published versions in these six low-resource languages, strong performance on probing tasks could indicate a high degree of cross-lingual knowledge transfer from English or other high-resource languages.\footnote{We exclude prefix probing from this experiment as it is unclear what the gold continuation would be.} Notably, we observe that models recall information even for the newly translated texts. GPT-4o, for instance, correctly identifies author and book title 69.4\% of the time and guesses masked entities with 6.3\% accuracy, suggesting that LLMs can, to some extent, access memorized knowledge across languages, even without direct exposure to these specific translations during pretraining \cite{yao-etal-2024-data, goldman2025eclekticnovelchallengeset}.\footnote{Although some models may be trained on machine translations, we see the same trend with OLMo, whose training data is public and can be inspected.}


\noindent\textbf{Third, we explore the robustness of memorization in cross-modal and quantized settings.} Our findings reveal that LLMs can recall memorized content even when prompted via different modalities, such as audio (GPT-4o-Audio achieves up to 75.5\% accuracy in direct probing; Qwen-Omni reaches 20.6\%). Furthermore, model quantization  impacts performance; for instance, LLaMA-3.1-70B shows up to a 25\% drop in accuracy with 8-bit quantization, a more substantial decrease than with 4-bit quantization, which contrasts with some previous findings \cite{marchisio-etal-2024-quantization, kurtic2025givebf16deathaccuracyperformance}.


\noindent\textbf{Contributions:} We introduce \dataname{}, a dataset featuring \textbf{31,540} aligned book excerpts across \textbf{10} languages. Using this dataset, we conduct three probing experiments to assess the extent of memorization by LLMs in English versus other languages, and to investigate how this memorized knowledge transfers across language boundaries. We are releasing our data and codebase to spur future research on multilingual memorization in LLMs.

%% file: sections/2-data-collection.tex
\section{Constructing OWL\,\raisebox{-0.2em}{\includegraphics[height=1.5em]{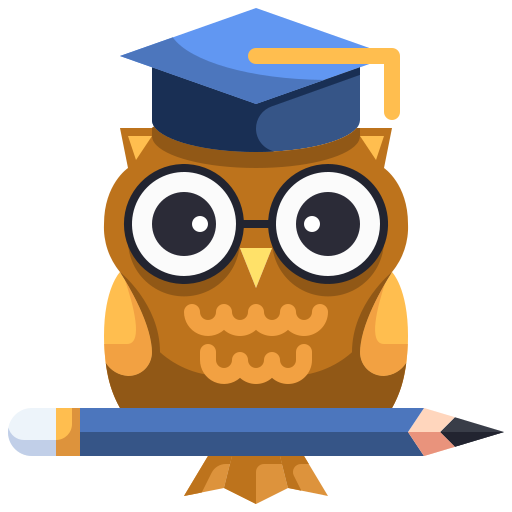}}}

\label{sec:data_collection}

\begin{table*}[t]
\centering
\footnotesize
\resizebox{\textwidth}{!}{%
\begin{tabular}{llll}
\toprule
\textsc{Passage Type} & \textsc{Perturbation} & \textsc{Experiment} & \textsc{English Example} \\
\midrule
\multirow{4}{*}{\parbox{1.8cm}{\centering \textsc{\faUser\ w/ character}}} 
 & \textsc{Standard} & DP + PP & ``Of course if Tom was home he'd put it right in a moment,''\\
 & \textsc{Masked} & DP + NC & ``Of course if \texttt{[MASK]} was home he'd put it right in a moment,''\\
 & \textsc{Shuffled} & DP & ``in he'd home Tom a if was of put it moment right course,'' \\
 & \textsc{Masked + Shuffled} & DP + NC & ``in he'd home \texttt{[MASK]} a if was of put it moment right course,'' \\
\midrule 
\multirow{2}{*}{\parbox{1.8cm}{\centering \textsc{\faUserSlash\ w/o character}}} 
 & \textsc{Standard} & DP &``No. Don't come up to me until you see me among a lot of people...'' \\
 & \textsc{Shuffled}& DP & ``Just me a you see at don't me. of me." people. Don't keep up...'' \\ %
\bottomrule
\end{tabular}
}
\caption{Examples of perturbations used in the ablation experiments. 
 \textbf{Experiment} indicates the evaluation setup the task appears in: DP = Direct Probe, PP = Prefix Probe, NC = Name Cloze.
 \textbf{English Example} shows a representative passage for each condition.}
\label{tab:perturbation-examples} 
\end{table*}

\begin{table*}[t]
\centering
\footnotesize
\resizebox{\textwidth}{!}{%
\begin{tabular}{l|rrrrrr|rrrrrr}
\toprule
 & \multicolumn{6}{c|}{\textsc{Original}} & \multicolumn{6}{c}{\textsc{No Named Characters}} \\
\textbf{Group} & \textbf{Count} & \textbf{Mean} & \textbf{Median} & \textbf{Min} & \textbf{Max} & \textbf{Stdev} 
& \textbf{Count} & \textbf{Mean} & \textbf{Median} & \textbf{Min} & \textbf{Max} & \textbf{Stdev} \\
\midrule
English       & 1594 & 64.90  & 49.0  & 18 & 429  & 47.75  & 1560  & 59.03  & 46.0  & 18  & 325  & 40.08  \\
Translations  & 4782 & 63.17  & 48.0  & 10 & 523  & 49.83  & 4680  & 57.67  & 45.0  & 10  & 430  & 43.01  \\
Cross-lingual  & 9564 & 78.91  & 60.0  & 11 & 642  & 59.98  & 9360  & 71.73  & 56.0  & 9   & 507  & 50.56  \\
\bottomrule
\end{tabular}

}
\caption{\label{tab:updated_token_stats}Token distribution in each passage type, calculated with OpenAI's \texttt{tiktoken} library (\texttt{o200k\_base}).}
\end{table*}

We design \dataname as a testbed for memorization as well as cross-lingual knowledge transfer in LLMs. The dataset has three main components: (1) excerpts from novels originally written in English (\textit{en}), (2) their official translations into Spanish (\textit{es}), Turkish (\textit{tr}), and Vietnamese (\textit{vi}), and (3) new machine translations into six low-resource languages, specifically Sesotho (\textit{st}), Yoruba (\textit{yo}), Setswana (\textit{tn}), Tahitian (\textit{ty}), Maithili (\textit{mai}), and Malagasy (\textit{mg}), for which official translations are not available. Additionally, we augment the data with audio files of the English excerpts to explore how models perform across modalities (text vs. audio).
Overall, we collect 3,154 English passages (1,594 passages with and 1,560 passages without named characters). Each passage is then aligned with its semantic equivalents in nine other languages and English audio, yielding a total of \textbf{31,540 text passages} and \textbf{7,950 audio excerpts} across the dataset. We construct the dataset in six main steps (\autoref{fig:pipeline}), as listed below:

\paragraph{1. Curating books}
We collect English novels that are also officially translated into Spanish, Turkish, and Vietnamese.\footnote{We selected these languages because they represent distinct morphological and syntactic typologies: Spanish is fusional, Turkish agglutinative, and Vietnamese analytic.}
We source public-domain books from Project Gutenberg \citep{stroube2003literary} and purchase copyrighted texts online. Overall, we collect \textbf{20 books}, with 10 public-domain and 10 copyrighted books (see \autoref{tab:book_stats}).

\paragraph{2. Tagging named characters}
Since the name cloze task (\S\ref{sec:nct}) requires test samples to have at least one character name, we tag these names by applying Stanza~\cite{qi2020stanzapythonnaturallanguage} to each sentence in the collected books.

\paragraph{3. Aligning multilingual paragraphs} To ensure fair comparison across languages, we align English passages to their official translations in Spanish, Vietnamese, and Turkish by translating non-English books into English using GPT-4o\footnote{We use gpt-4o-2024-05-13 with temperature=0.3 and max\_tokens=4000; refer to \autoref{fig:translation-prompt} for details.} and applying the Par3 aligner \cite{thai2022exploringdocumentlevelliterarymachine}.

\paragraph{4. Filtering \& quality control} To filter out any misaligned passages, we apply a length filter\footnote{We use an asymmetric length filter: drop if the English passage is >3x any non-English by characters ($|x_{\text{en}}|>3|x_\ell|$).} and BLEU filter using SacreBLEU~\cite{post-2018-call} with add-one smoothing.\footnote{We filter out any alignment that does not meet the threshold of 5.0 BLEU score, following \citet{thai2022exploringdocumentlevelliterarymachine}} Finally, we manually verify all alignments, removing misaligned passages or those with more than one unique character name (\autoref{fig:labelstudio}).
We compile two sets of passages: (1) a set containing exactly one unique character name\footnote{This is our main experimental set; for these passages, we allow multiple mentions of the same character's name.} that is used for all our tasks, and (2) a set of comparable size that does not have any character name for the direct probing and prefix probing task (\S\ref{sec:direct-probing}). The two sets have similar average lengths: 64.90 tokens for passages with a character name and 59.03 for those without
(\autoref{tab:updated_token_stats}).\footnote{Unless otherwise mentioned, ``tokens" refer to those calculated with \texttt{tiktoken} library (\texttt{o200k\_base})}  For each set of passages, we sample at most 100 passages per book to include in the final dataset.\footnote{We sample passages with at least 40 BPE tokens. View word count distribution in \autoref{fig:wordcountdis}.} 

To balance the distribution of character mentions, we apply stratified sampling to all passages containing a character name. The final dataset for each language consists of 3,154 passages: 1,594 with character names and 1,560 without.\footnote{This addresses a bias from \citet{chang2023speakmemoryarchaeologybooks}, where overrepresentation of common names (e.g., Alice) likely inflated model accuracy.}


\noindent\paragraph{5. Machine translation into new languages} To explore cross-lingual knowledge transfer, we select six languages with \textit{no prior translations} of the books in our dataset to ensure that they have not been encountered during the training: Sesotho (\textit{st}), Yoruba (\textit{yo}), Setswana (\textit{tn}), Tahitian (\textit{ty}), Maithili (\textit{mai}), and Malagasy (\textit{mg}).\footnote{To confirm no existing translations, we search Google, Amazon Books, OpenLibrary, and Goodreads for each book in the target language and find none.} We use Microsoft Translator\footnote{\url{https://www.microsoft.com/en-us/translator/}. We used Microsoft Translator instead of LLMs to avoid potential bias and because LLMs generally underperform traditional machine translation on low-resource languages due to data limitations \cite{robinson-etal-2023-chatgpt}.} to translate passages from English into each of the unseen languages.\footnote{We recognize that Microsoft Translator may not produce perfect translations; therefore, the results presented in this paper represent a lower bound of the cross-lingual performance.} We will be referring to this subset of data as \textit{unseen translations}.

\noindent\paragraph{6. Creating audio data} 

To evaluate cross-modal knowledge transfer, we convert passages containing character names into high-fidelity, lossless audio waveforms using Kokoro-82M \cite{hexgrad_2025}, a neural text-to-speech (TTS) model chosen for its low-distortion rendering of prosody and phonetics. The resulting audio corpus preserves the linguistic content of each passage while enabling direct comparison between text and speech-based representations.\footnote{Kokoro-82B currently ranks as the top-performing TTS model on TTS Spaces Arena \cite{tts-arena-v2}. A manual review of 50 samples revealed no errors.}
We convert the entire passage into audio for all tasks. For prefix probing (\S\ref{subsec:prefix-probing}), we convert only the first half of the passages.

\paragraph{\faBook\ Why literary data?}
We select literary work as it is likely present in pretraining corpora. All our titles are available on LibGen (allegedly used to train LLaMA models),\footnote{See \href{https://fingfx.thomsonreuters.com/gfx/legaldocs/lbvgjdkdopq/META\%20COPYRIGHT\%20LAWSUIT\%20libgen.pdf}{legal brief}.} and our non-copyrighted books are on Project Gutenberg (used for OLMo training \cite{olmo20242}). Furthermore, literary data is rich in the character names necessary for the name-cloze task.

%% file: sections/3-experiments.tex
\section{Experiments}
\label{sec:experiments}

We propose three probing experiments on \dataname to assess memorization as well as cross-lingual knowledge transfer: (1) direct probing (DP, \S\ref{sec:direct-probing}), where the model identifies the book's title and author; (2) name cloze (NC, \S\ref{sec:nct}), where the model fills in a masked character name; and (3) prefix probing (PP, \S\ref{subsec:prefix-probing}), where the model generates a continuation from a given prefix. We further extend these experiments to test the effect of quantization on memorization (\S\ref{subsec:quant-abl}), and probe cross-modal recall using audio input (\S\ref{subsec:audio-abl}).
These tasks reflect different degrees of memorization, ranging from basic retrieval of learned information (direct probing) to precise reconstruction of acquired content (prefix probing). 

\paragraph{\faFile\ Test data} Unless specifically mentioned, we run all experiments on the following data containing one unique character name: \textbf{(1) original English data} (to establish model recall of data, which was likely seen during pretraining), \textbf{(2) official translations} (to provide a baseline for model's performance on high-resource languages other than English, which could be encountered during pretraing), \textbf{(3) unseen translations} to measure cross-lingual knowledge transfer, and \textbf{(4) English audio data} (to compare performance on audio and textual content). We also include an additional experiment on newly published books to estimate performance by chance.

\subsection{{\colorbox{SkyBlue!80}{\faFlask\ Experiment 1:}}  Direct Probing}
\label{sec:direct-probing}

\paragraph{\faClipboardList\ Task:} In direct probing, the model identifies the \texttt{title} and \texttt{author} of a book passage \cite{karamolegkou-etal-2023-copyright}. This task reflects more passive knowledge, as it primarily tests the model's ability to recognize and link textual and audio cues to learned metadata rather than requiring the model to recall the exact wording of the passage (see \autoref{fig:direct-probing} for prompt). In the cross-modal setup, we provide the audio of the passage.

\paragraph{\faChartLine\ Metric:} We measure accuracy by comparing predicted \texttt{(author, title)} pairs against ground truth, allowing for minor formatting or diacritic differences.\footnote{We normalize special characters and apply fuzzy match with a Levenshtein similarity threshold (0.9 for DP and 0.7 for NC, which we establish by analyzing a subset of our data).} A prediction is considered correct if the model identifies the correct author and book title (either in English or the passage's language). For cross-lingual experiments, we prompt the model to respond in English. 

\paragraph{\faEraser\ Ablations:} To measure the robustness of model performance, we introduce three additional variations on the task (see \autoref{tab:perturbation-examples}):

\textit{Shuffled passages:} To pinpoint the role of word order and syntax in knowledge recall, we randomly shuffle the words within each passage. This shuffle disrupts the syntactic and semantic coherence of the text while preserving its lexical content, allowing us to test whether the recall depends on the sequential structure of the input.  

\textit{Masked passages:} For consistency across tasks, we use the same passages as in the name cloze task (\S\ref{sec:nct}), each containing a single character name. Here, we replace that name with \texttt{[MASK]} to determine how much it contributes to the recall, albeit at the cost of disrupting the original text.

\textit{No character names:} We also include a separate set of passages that naturally contain no character names and thus remain intact. To facilitate a fair comparison with masked passages, we ensure that both sets have similar length distributions.

\subsection{{\colorbox{SkyBlue!80}{\faFlask\ Experiment 2:}}  Name Cloze}
\label{sec:nct}

\paragraph{\faClipboardList\ Task:}  In the name cloze task, we reuse the same passages from \S\ref{sec:direct-probing}, each containing exactly one character name, and replace that name with \texttt{[MASK]} token to test recall \citep{chang2023speakmemoryarchaeologybooks}.\footnote{Unlike \citet{chang2023speakmemoryarchaeologybooks}, we do not restrict passages to a single occurrence of the character name or limit the passage length to allow for more realistic text usage and analysis of passage-length effects.} Strong performance on this task likely indicates memorization of that passage, especially since character names tend to be high-surprisal tokens~\cite{ravichander2025halogenfantasticllmhallucinations}. In the cross-modal setup, we provide the English audio of the passage. 

\paragraph{\faChartLine\ Metric:} We evaluate task accuracy using exact match.\footnote{Exact match is applied after normalizing both predicted and ground-truth names with the Unidecode library to remove formatting and diacritic variations.} Ground-truth named characters are extracted directly from the original passages, and a prediction is correct only if it matches the normalized ground truth (either in English or in the language of the passage). For cross-lingual experiments, we prompt the model to respond in English.  
\paragraph{\faEraser\ Ablation:}  We further test the robustness of models by shuffling the words within each passage, as previously done in \S\ref{sec:direct-probing}, to understand the effect of sequential token order and syntax. Specifically, we want to understand whether the model performance depends on the token sequence and/or the position of the \texttt{[MASK]} token. 

\subsection{{\colorbox{SkyBlue!80}{\faFlask\ Experiment 3:}}  Prefix Probing}
\label{subsec:prefix-probing}

\paragraph{\faClipboardList\ Task:} The prefix probing task evaluates whether a model, when given the first half (prefix) of a passage, can reproduce the second half (continuation) \cite{carlini2021extractingtrainingdatalarge}. This setup draws on the fact that accurate predictions are unlikely without prior exposure to the full passage during pretraining. In the cross-modal setup, we provide the English audio of the first half of the passage.

\paragraph{\faChartLine\ Metric:} To measure the model's ability to replicate a passage's continuation, we report ChrF++ \citep{popovic-2015-chrf}, which assesses lexical and semantic similarity between the model's output and the ground-truth continuation. 

\begin{figure*}[t]
    \centering
    \includegraphics[width=\textwidth]{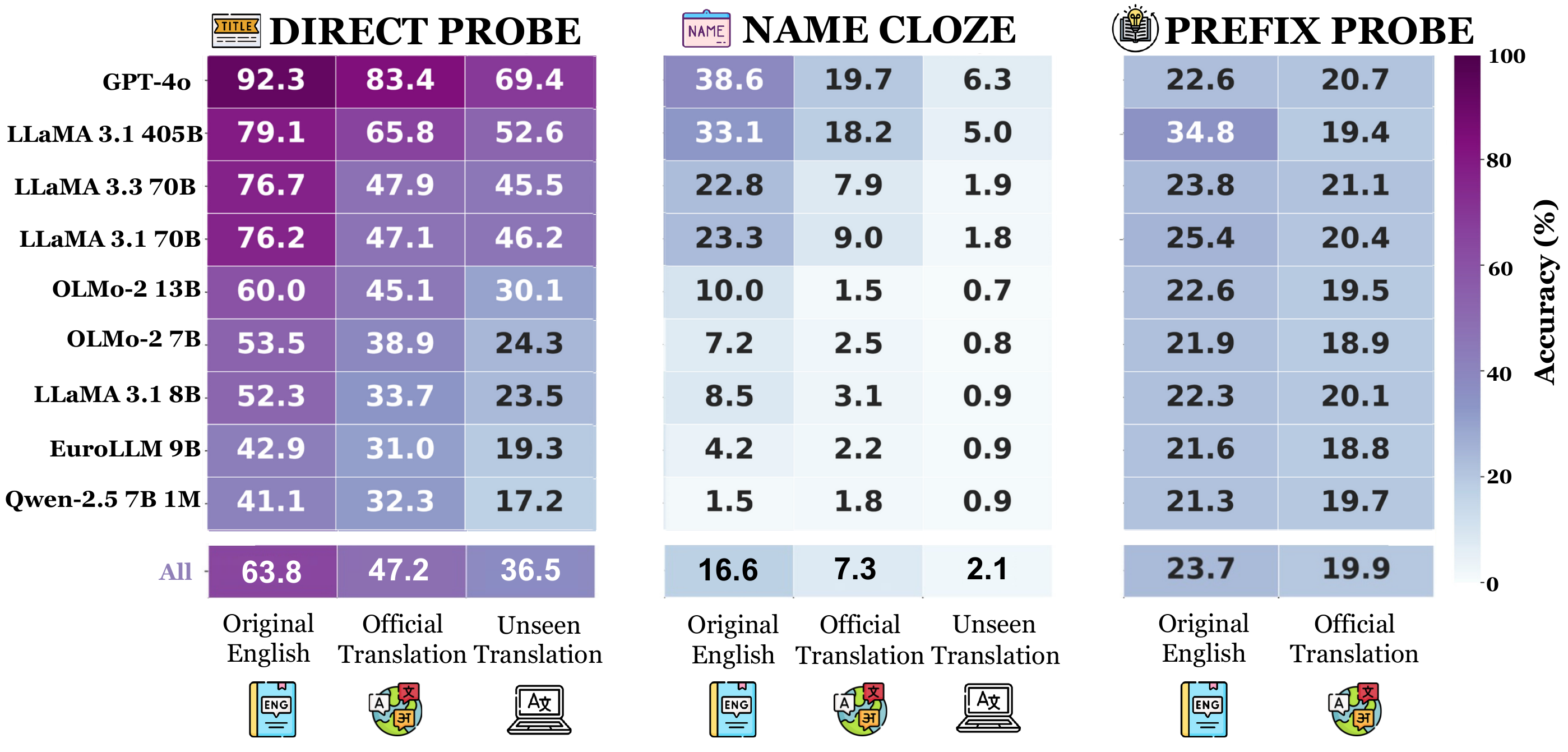}
    \caption{\textbf{Overall performance:} GPT-4o consistently outperforms other models in probing tasks, followed by LLaMA 405B. Direct Probing (DP; reported for passages with character names) and Prefix Probing (PP) use unmasked passages, while the Name Cloze Task (NCT) uses masked ones with named characters removed. PP performance is measured with ChrF++. PP performance on unseen languages is not reported as it is unclear what the gold continuation should be.}
    \label{fig:dp_nct_pp_heatmap}
\end{figure*}

\subsection{\faCubes\ Quantization ablation}
\label{subsec:quant-abl}

To assess potential information loss due to reduced parameter precision from quantization, we replicate all experiments and ablations on LLaMA models using GPTQ-int4 (\textsc{W4A16}) and GPTQ-int8 (\textsc{W8A16}) methods \cite{frantar2023optq}, where W\textit{x}A\textit{y} denotes the level of quantization for weights (W) and activations (A).

\subsection{\faHeadphones\ Audio ablation}
\label{subsec:audio-abl}

To compare performance on audio versus text, we extend our analysis to audio content, adapting three core experiments: direct probing, name cloze task, and prefix probing. Text-specific ablations were excluded. Due to superior text-to-speech (TTS) model quality, all audio experiments were limited to English, with models receiving textual instructions and providing textual responses.

\subsection{\faCogs\ Models}
For all tasks, we test a diverse set of open-weight and closed-source models, including Qwen2.5-1M \cite{qwen2.5, Qwen2.5-Omni}, LLaMA-3.1-8B, 70B, 405B and LLaMA-3.3-70B \cite{Dubey2024TheL3}, OLMo-2-7B and 2-13-B \cite{olmo20242}, EuroLLM \cite{MARTINS202553}, as well as GPT-4o~\cite{openai2024gpt4ocard}.\footnote{We use vLLM \cite{kwon2023efficientmemorymanagementlarge} for inference from open-weights models, with the exception of LlaMA-3.1-405B-instruct, which is run using OpenRouter API due to its size. For all models, we set the temperature to 0 and max\_tokens to 100.} For audio experiments, we use GPT-4o-audio and Qwen2.5-Omni-7B \cite{Qwen2.5-Omni}. In addition to full-precision models, we also run our experiments on the quantized versions of LLama-3.1-70B-Instruct and Llama-3.1-8B-Instruct.\footnote{Quantized models are obtained from \href{https://huggingface.co/collections/neuralmagic/int8-llms-for-vllm-668ec32c049dca0369816415}{NeuralMagic}.} See \autoref{api-pricing} for details.

%% file: sections/4-analysis.tex
\section{Results}

\label{sec:analysis}

In this section, we present the results of our experiments. Overall, our results show that LLMs can, to varying degrees, recognize (and in some cases reproduce) book content when presented in different forms, such as the original English text, official translations, new machine translations, and even audio. While perturbations (e.g., shuffling) do reduce accuracy, the resulting performance is still above random. Finally, the presence of a character's name proves to be a strong signal that facilitates recall.

\paragraph{LLMs can recognize official translations} Models can recognize passages from English novels achieving 63.8\% accuracy on average, with GPT-4o reaching 92.3\% (\autoref{fig:dp_nct_pp_heatmap}). Although this accuracy drops for official translations, it remains above random at 47.2\% on average (83.4\% for GPT-4o).
This recall also extends to more challenging tasks such as name cloze, albeit with reduced accuracy (e.g., GPT-4o scores 38.6\% for English versus 19.7\% for translations; see \autoref{tab:custom_error_types} for common errors). 
Notably, performance scales with model size. In the name cloze task for English texts, accuracy rises from 8.5\% with LLaMA-3.1-8B to 33.1\% with LLaMA-3.1-405B. These results indicate memorization, particularly in comparison with the performance on 2024 books (\autoref{tab:2024}), where the accuracy is close to zero, likely because the content was not seen during training. Finally, prefix probing results suggest the models struggle with verbatim recall, as their chrF++ scores are only marginally higher than those for the 2024 books.

\paragraph{Cross-lingual access to memorized knowledge} Having established that models can recognize English excerpts and their official translations, we next test whether they could also recognize newly produced machine translations in six low-resource languages. Although the overall accuracy drops, the models can still identify the books (36.5\% average on the direct probe; \autoref{fig:dp_nct_pp_heatmap}) and, to a lesser extent, recall a masked character's name (2.1\% average on name cloze). This performance varies by language and model (\autoref{fig:xling}). For instance, on the direct probe for Sesotho, GPT-4o achieves 76.9\% accuracy, while Qwen-2.5-7B-1M scores over 18\%. Even for Maithili, the lowest-performing language, GPT-4o still achieves 66.5\% accuracy, with LLaMA-3.1-405B close behind at 46.7\%. 

The name cloze results are much lower but still above zero, with the highest score being 10.5\% on Maithili by GPT-4o.\footnote{This performance varies significantly by book, for example, GPT-4o achieves an average accuracy of 33.3\% for "Alice's Adventures in Wonderland" and 19.2\% for "1984" when tested on unseen translations.} Interestingly, even OLMo shows a non-zero performance, despite being reportedly trained only on English data \cite{olmo20242}, with its highest score being 44.1\% on the Yoruba direct probe.\footnote{Note that while OLMo's authors specifically filtered their training data for English, it's very likely that some non-English text remained.} This all suggests that some meaningful amount of cross-lingual transfer can happen even when the target languages are underrepresented in the pre-training data.\footnote{\href{https://chatgptiseatingtheworld.com/wp-content/uploads/2025/01/Unredacted-Reply-of-Plaintiffs-1.pdf}{Recent court documents} show that LlaMA models might have been trained on LibGen book data, which implies the possibility that other frontier models are also trained on this data (100\% of our English and official translation books can be found in LibGen). Under these conditions, it is reasonable that these models have boosted cross-lingual performance without explicit translation supervision.} \footnote{It is worth noting that the character's name is a strong signal, a point we discuss later in more detail. Removing the name from a passage lowers accuracy on the direct probing task, although performance doesn't fall to zero (\autoref{fig:heatmap-dp-clm}).}

\begin{figure}[ht]
  \centering
  \includegraphics[width=\columnwidth]{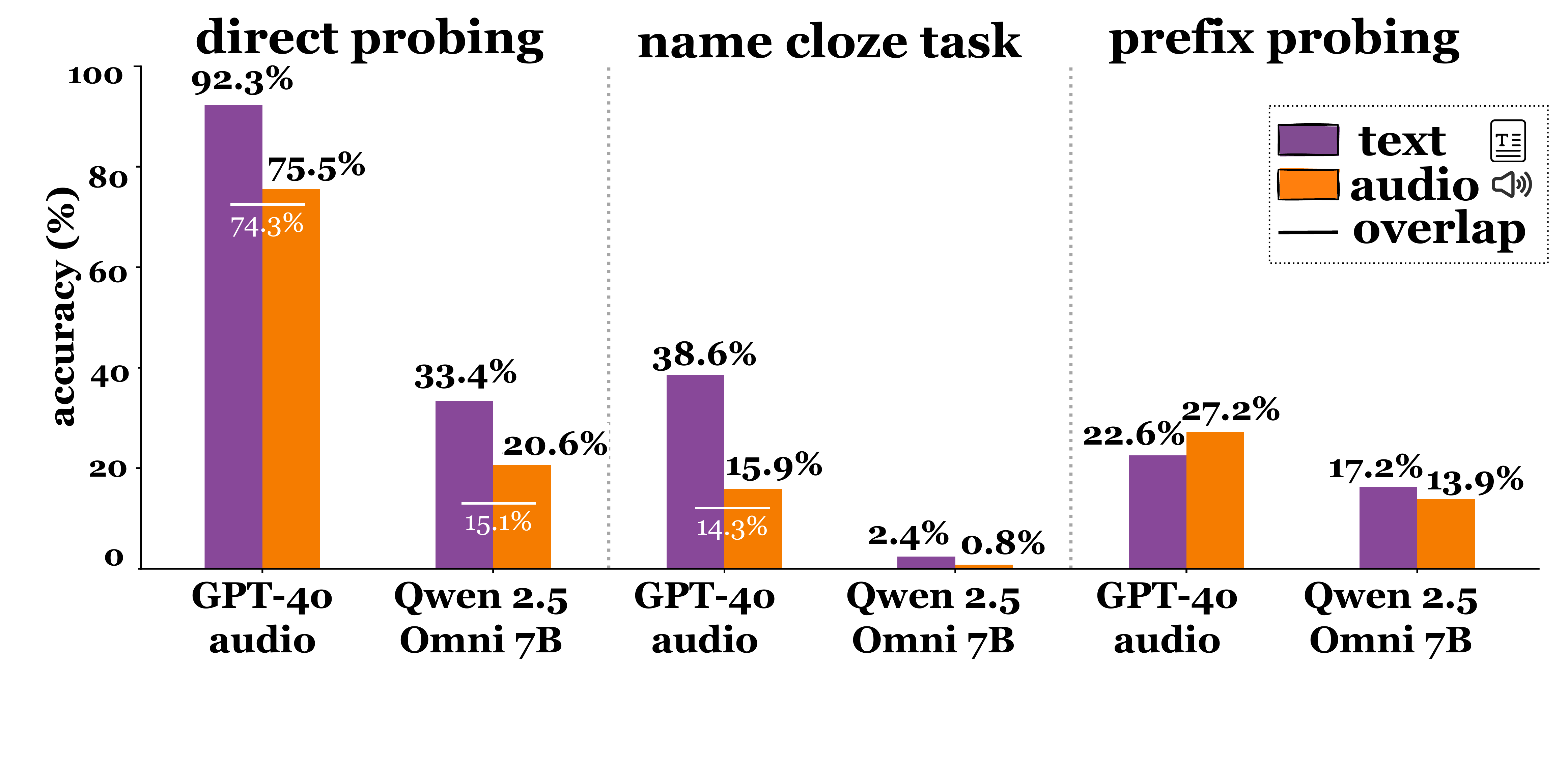}
  \caption[\textbf{Audio vs.\ Text} Comparison]
  {%
    \textbf{Audio vs.\ Text} accuracy on English passages with a character name. GPT-4o-audio exhibits substantial performance across all tasks and modalities. The overlap line denotes the percentage of passages answered correctly in both modalities.
  }
  \label{fig:audiovtext}
\end{figure}

\paragraph{LLMs can recall knowledge when probed in a different modality} 
Both Qwen-Omni and GPT-4o-Audio show some ability to recognize the book when prompted with an audio excerpt (\autoref{fig:audiovtext}, \autoref{fig:audiovtext_overlap}). Specifically, GPT-4o-Audio achieves up to 75.5\% accuracy on the direct probing task, while Qwen-Omni reaches 20.6\% on the same task. Although overall performance is lower on the audio version of the name cloze task, GPT-4o-Audio still reaches up to 15.9\%. In contrast, Qwen-Omni struggles with this task, scoring only 0.8\%. These findings suggest that LLMs could potentially recall information across modalities.\footnote{It is possible the models were trained on official audiobooks. However, our audio files were created using a text-to-speech model and thus differ significantly in their acoustic properties from human narrations.}

\begin{figure}[t]
  \centering
  \resizebox{0.99\columnwidth}{!}{%
    \includegraphics{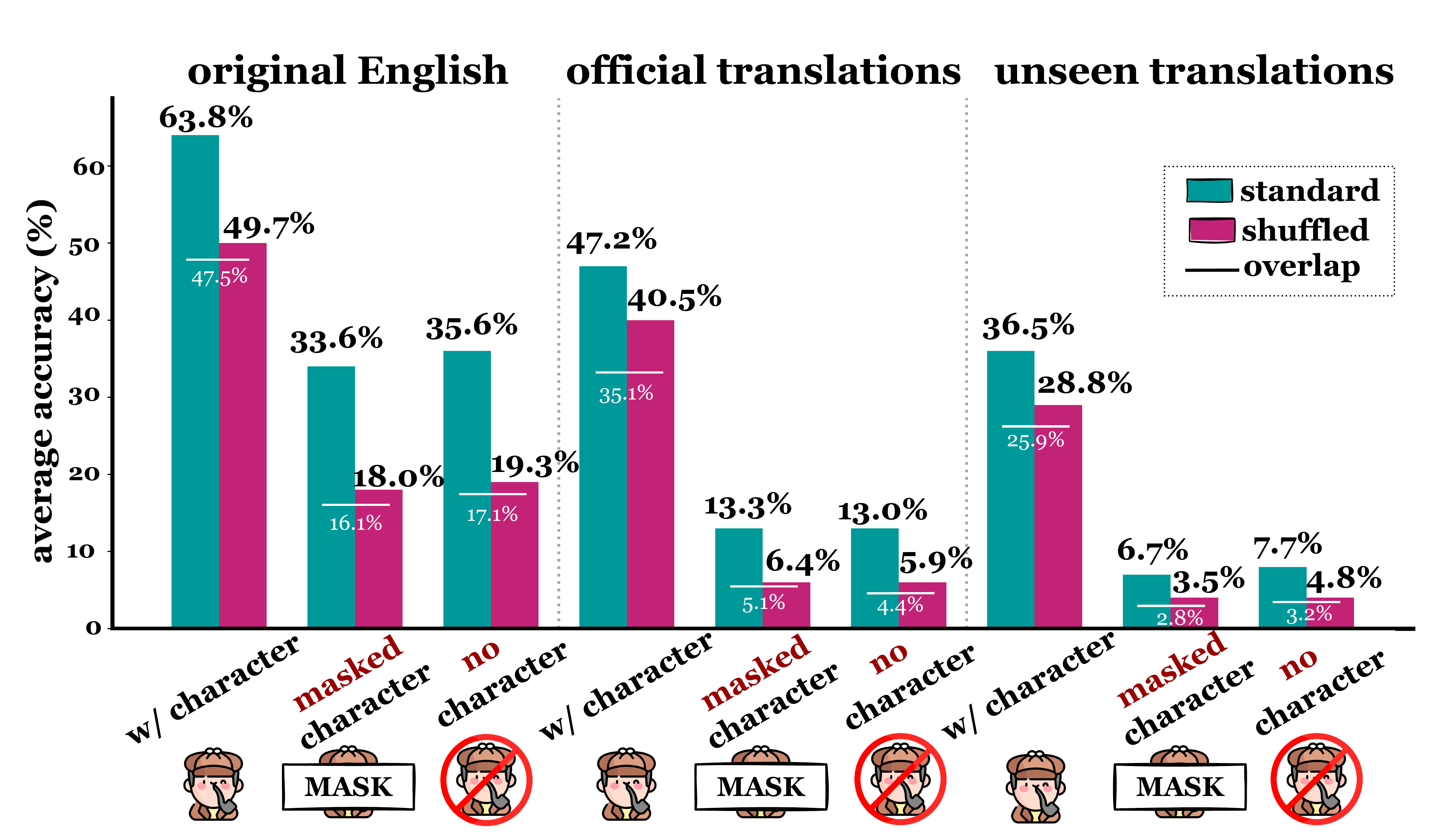}%
  }
  \caption{\label{fig:dp_shuffled} \textbf{Direct probing:} Average accuracy across models for shuffled versus standard text inputs. Accuracy decreases from standard to shuffled inputs across all perturbations and language settings, with non-trivial shuffled accuracy on English and official translations.}
\end{figure}

\begin{figure}[t]
  \centering
  \resizebox{0.8\columnwidth}{!}{%
    \includegraphics{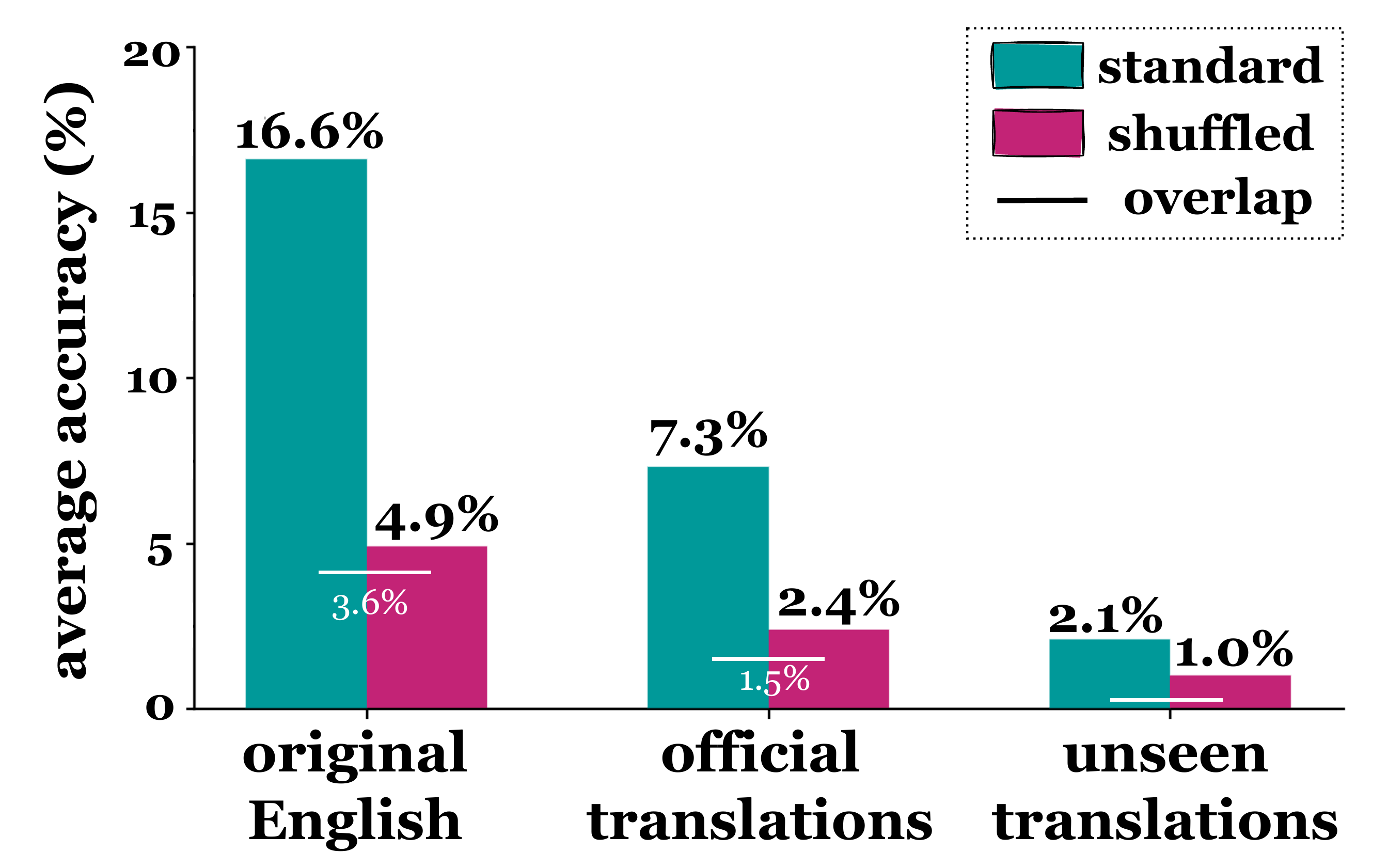}%
  }
  \caption{\label{fig:nct_shuffled} \textbf{Name cloze:} Unshuffled inputs outperform shuffled inputs across all language settings, with non-trivial accuracy on English and official translations.}
\end{figure}

\paragraph{Shuffling inputs partially reduces direct probing and name cloze accuracy}
\autoref{fig:dp_shuffled} shows that shuffling the input texts, which represents minor perturbations such as phrase reordering or lexical edits, causes a noticeable, but not drastic, drop in direct probing accuracy. Specifically, direct probing performance declines most on English passages (up to 16.3\% for ``no character'' setup), in contrast to 3-8\% for both translation categories across all excerpt types; likely because the accuracies on these are already lower. Similarly, in the name cloze (\autoref{fig:nct_shuffled}) the gap between standard and shuffled performance can be as low as 1.1\% for unseen translation and as high as 11.7\% for English texts. These moderate drops indicate that superficial rewordings do have an effect, though only a modest one.

\paragraph{Direct probing consistently outperforms name cloze-style queries}
Direct probing outperforms name cloze queries across all models and languages (\autoref{fig:dp_nct_pp_heatmap}). For example, GPT-4o achieves 92.3\% accuracy on original English texts with direct probing, compared to only 38.6\% with name cloze. LLaMA 3.1 70B shows a similar gap (76\% vs. 22.8\%), as does EuroLLM 9B (38.7\% gap). This pattern holds in translations: GPT-4o scores 83.4\% (direct) vs. 19.7\% (cloze) on official translations, and 69.4\% vs. 6.3\% on unseen ones. The large performance gap reflects the difficulty of name cloze tasks, which likely conflict with the autoregressive nature of language models. In contrast, direct probing, where the model has to recall the title or author in a question-answering format, is more aligned with LLMs' strength.

\paragraph{Character names facilitate recall}
\autoref{fig:dp_shuffled} shows that models are noticeably better at the direct probing task, when the passage contains a character name: 63.8\% for English, 47.2\% for official translations, and 36.5\% for unseen ones. Masking the name sharply reduces accuracy to 33.6\%, 13.3\%, and 6.7\%, respectively. Accuracy under masking is similar to passages without named characters, especially in translations ($\leq$3\% difference).
The absence of named characters results in lower performance, suggesting that models often depend on lexical cues like names and locations to recognize the passages.

\begin{figure}[t]
    \centering
    \includegraphics[width=1\linewidth]{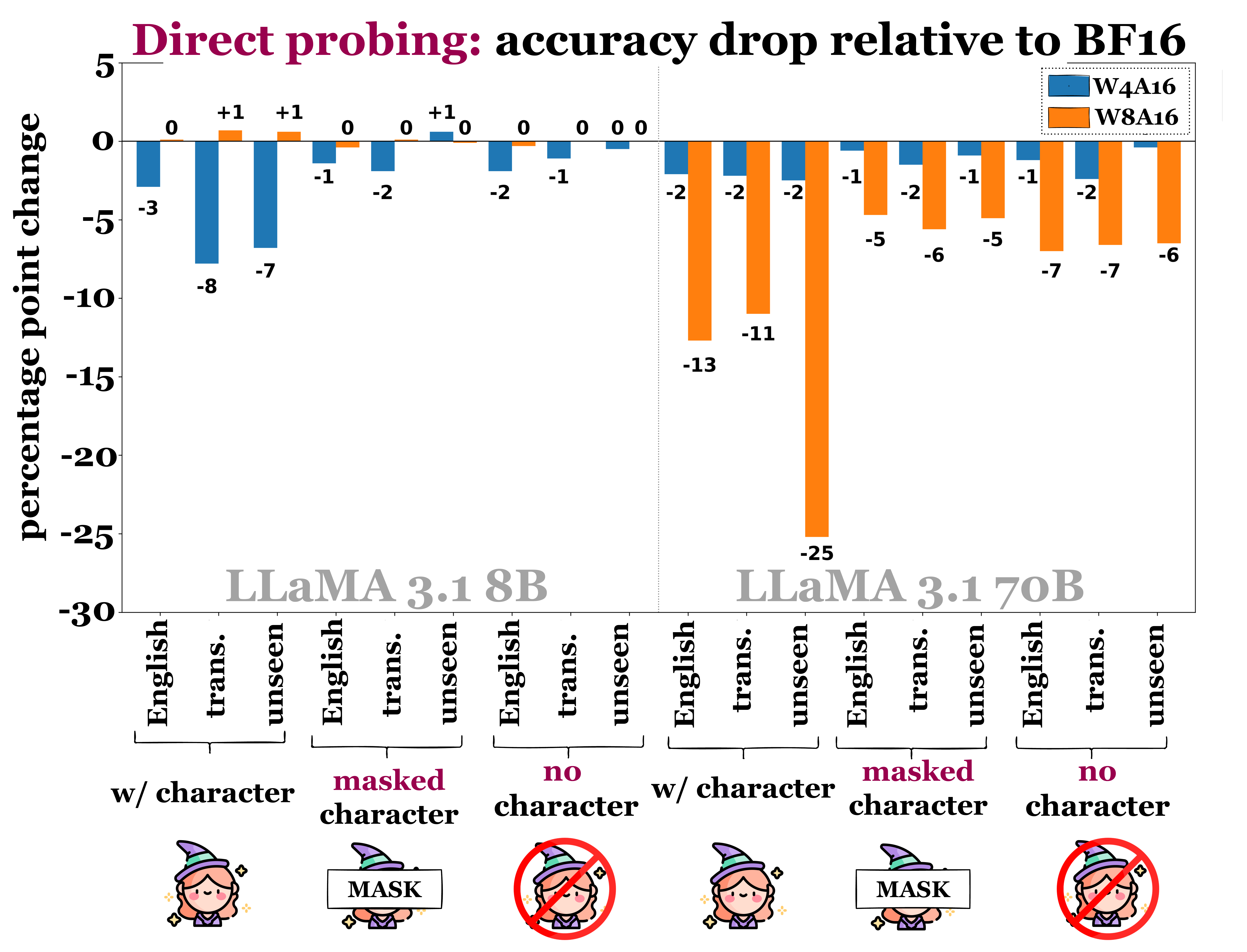}
    \caption{\textbf{Direct probing:} Percentage point drop in performance with respect to the performance of the BF16 baseline. We report drops for original English text ("English"), their official translations ("trans"), and unseen translations ("unseen"). The scores are reported across three conditions: (1) on passages containing a character name, (2) on passages where the name was masked, and (3) on passages without character name.
    }
    \label{fig:dp-quant}
\end{figure}

\paragraph{LLaMA-3.1-70B's performance degrades more under 8-bit than under 4-bit quantization} 
We test the performance of both LLaMA-3.1-70B and 8B under 4-bit and 8-bit quantization and compare it to the performance of these models in BF16 precision.\footnote{We report the performance drop as the difference in percentage points between the BF16 version and quantized models.}
While LLaMA-3.1-70B maintains relatively stable accuracy at 4-bit precision, it experiences notable performance drops when quantized to 8 bits. Specifically, we observe up to a 25\% decrease in direct probing on unseen translation settings (\autoref{fig:dp-quant}), along with smaller declines of 5.8\% in the English name cloze task and 1.4\% in prefix probing when the passage includes character names (\autoref{fig:pp-quant}). In contrast, the smaller LLaMA-3.1-8B behaves more predictably: its performance remains within 1\% of the BF16 baseline at 8-bit precision across both direct probing and name cloze tasks (\autoref{fig:nct-quant}), with noticeable degradation appearing only under 4-bit quantization. These results surprisingly contradict findings in \citet{kurtic2025givebf16deathaccuracyperformance} and \citet{marchisio-etal-2024-quantization}, who report a marginal drop for GPTQ-int8 but larger drops for GPTQ-int4.

\begin{figure}[t]
    \centering
    \includegraphics[width=0.8\linewidth]{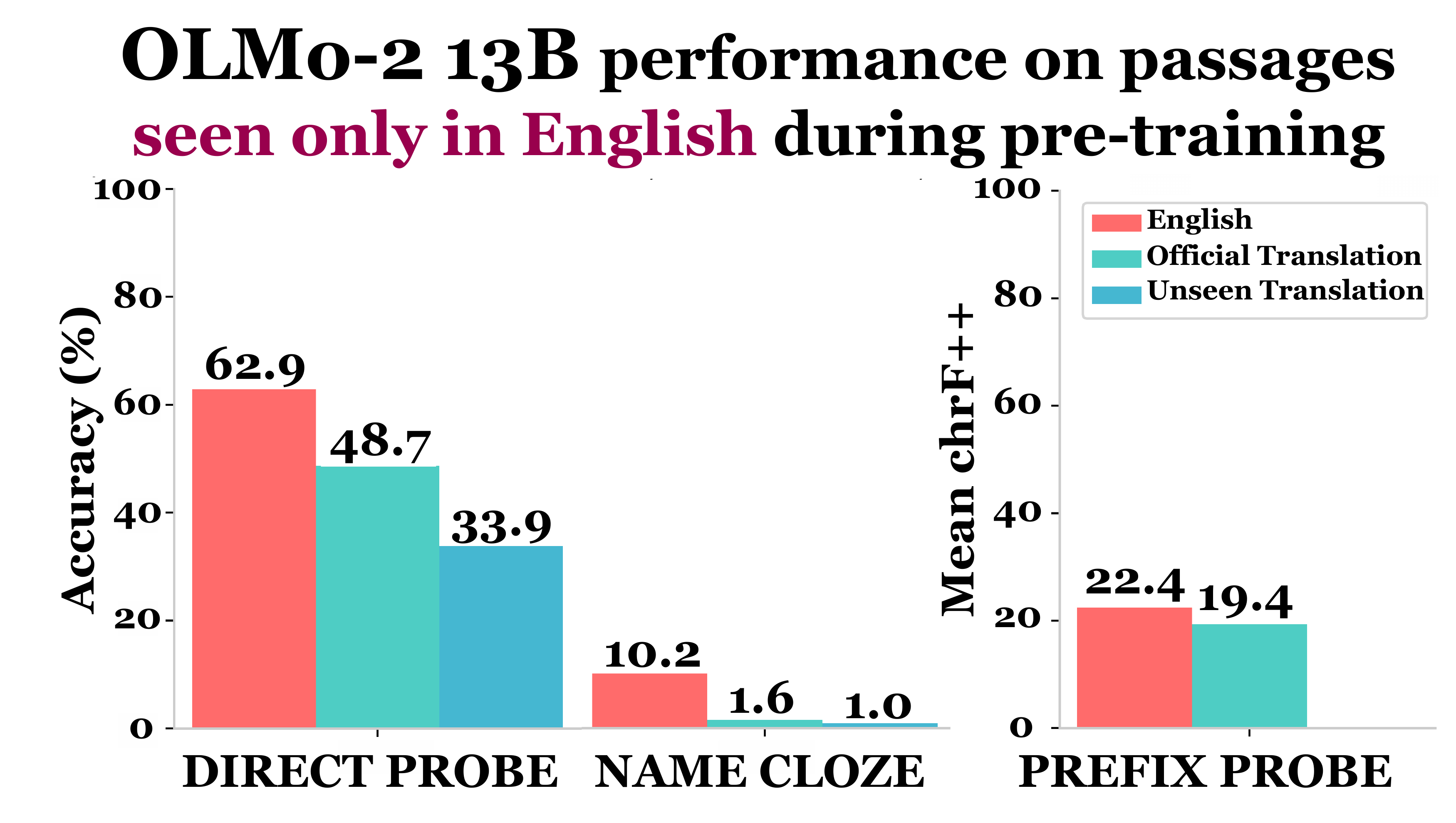}
    \caption{Accuracy of OLMo-2-13B on \emph{seen} passages identified in the training data in English but not in their translated versions. The model's accuracy on direct probing is considerable compared to its performance on name cloze and prefix probing.
    }
    \label{fig:olmo213-seen}
\end{figure}

\begin{figure}[t]
    \centering
    \includegraphics[width=0.8\linewidth]{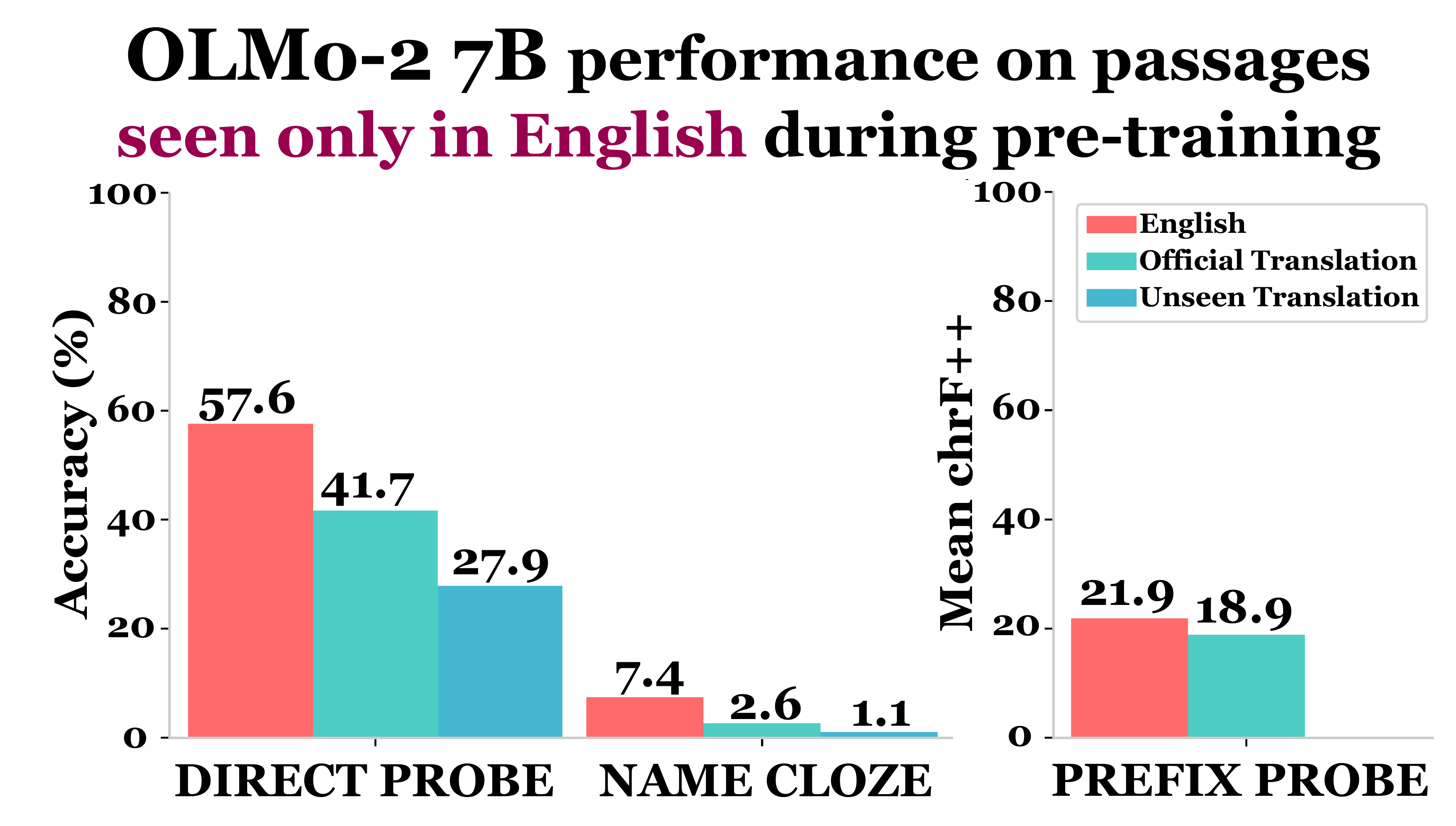}
    \caption{Accuracy of OLMo-2-7B on passages identified in the training data in English but not in their translated versions. Accuracy across tasks and languages is slightly lower than that of the 13B model.
    }
    \label{fig:olmo27-seen}
\end{figure}

\paragraph{OLMo's performance on passages seen during training}
\label{sec:olmo-check}
We identified a subset of \dataname{} for which the English passages are present in OLMo's training data, and we made sure that none of the corresponding translations were.\footnote{Passages were identified by querying the Infinigram API \cite{Liu2024InfiniGram} (\texttt{v4\_olmo-2-1124-13b-instruct\_llama}). See \autoref{tab:infinigram-non-ne} for details.} Even though the OLMo models had definitely seen these English passages during training, their performance on them is moderate ($\sim$62.9\%). However, the accuracy does not drop drastically on the direct probing task for both official and machine translations of those same passages (\autoref{fig:olmo213-seen}, \autoref{fig:olmo27-seen}). Unsurprisingly, the performance on the name cloze task is much lower, though it still likely remains above random.

\paragraph{Analysis of common errors} For
direct probing, models occasionally name correct authors but misidentify book titles around 10.61\%\footnote{A common error pattern involves models correctly attributing authorship to J.K. Rowling but specifying an incorrect book title from within the Harry Potter series.} of the time. 
More often, they return another popular book
(61\%) or abstain (39\%).
Abstention rate (responses like ``unknown,'' ``none,'' or empty strings) is notably high for EuroLLM (30.39\%). The main error for name cloze task is returning an incorrect character name (93\%), which is sometimes culturally relevant to the passage's language (e.g., Spanish names for Spanish text) or other characters from the same book (\autoref{tab:top_names}). Models also return pronouns (2\%), honorifics (3\%), abstain (0.1\%), or repeat the ``[MASK]'' token (0.7\%).
Across all tasks, Qwen models frequently generate ``broken text'' (a hodgepodge of languages) for 15.81\% of outputs. See \autoref{tab:custom_error_types} for more examples.

%% file: sections/5-related-work.tex
\section{Related Work}
\label{sec:rel_work}
\paragraph{Memorization in LLMs}
LLMs exhibit substantial memorization capabilities \citep{elangovan-etal-2021-memorization,Carlini2018TheSS,hartmann2023sokmemorizationgeneralpurposelarge,carlini2023quantifying}. Prior studies quantify memorization through verbatim recall \citep{carlini2021extractingtrainingdatalarge,carlini2023quantifying,lee-etal-2022-deduplicating}, passage origin identification \citep{chang2023speakmemoryarchaeologybooks,magar2022datacontaminationmemorizationexploitation}, improbable token prediction \citep{lee-etal-2022-deduplicating,radhakrishnan2019memorizationoverparameterizedautoencoders}, and membership inference attacks \citep{Carlini2021MembershipIA,Golchin2023TimeTI,song2019auditingdataprovenancetextgeneration,7958568,Asai2020XORQC,stoehr2024localizing}. The amount memorized depends on factors such as model scale, generation length, data point frequency, and context window size \citep{carlini2023quantifying,radhakrishnan2019memorizationoverparameterizedautoencoders,10.5555/3618408.3618510,Zhou2023QuantifyingAA,chen2024multiperspectiveanalysismemorizationlarge, razeghi-etal-2022-impact,lee-etal-2022-deduplicating,kandpal2022deduplicating,carlini2023quantifying}. Early probing experiments, which are largely monolingual and cloze-style \citep{tirumala2022memorization,chang2023speakmemoryarchaeologybooks}, have since been complemented by theoretical work showing that rare memorized outliers can steer the model's learning trajectory \citep{allen-zhu2024physics}. \citet{prashanth2025recite} model memorization as three regimes -- recitation, reconstruction, recollection -- and fit a predictive model. \citet{DBLP:journals/corr/abs-2506-12321} study memorization dynamics in Pythia and find that prefix perturbations reduce recall; our perturbations overlap in spirit, and we defer full replication to future work.

\paragraph{Cross-lingual knowledge transfer}
Cross-lingual knowledge transfer enables LLMs to recall information seen in one language when queried in another through shared multilingual representations \citep{asai-etal-2021-xor, jiang-etal-2020-x, limkonchotiwat-etal-2022-cl, mittal-etal-2023-mokb6, litschko-etal-2024-improving}. Research in both multimodal \citep{elliott-etal-2016-multimodal, baltrusaitis2017multimodal} and multilingual settings \citep{hessel-lee-2020-multimodal} has shown that models can achieve high performance by exploiting shallow or dataset-specific cues.
Our work is most relevant to \citet{goldman2025eclekticnovelchallengeset}, who measure cross-lingual transfer by analyzing the presence or absence of Wikipedia entries across languages and evaluating LLMs on this data.

%% file: sections/6-discussions_conclusion.tex
\section{Conclusion}
In this study, we demonstrate that LLMs exhibit some degree of multilingual and cross-lingual memorization through probing experiments on aligned book excerpts across ten languages. We discover that character names are a strong signal for recalling the information. We also find that perturbations, such as word shuffling, prompting in audio format, and masking character names, noticeably but not drastically reduce performance. We release our data and code to spur further research on cross-lingual generalization and LLM memorization. 

%% file: sections/7-limitations.tex
\section*{Limitations}

\paragraph{Material scope} We study memorization using best-selling books, which might not reflect the full diversity of copyrighted materials. Future work should explore additional underrepresented languages and lesser-known texts. 

\paragraph{Popularity versus performance} Models might have higher performance on excerpts that appear frequently in the pretraining data \citep{carlini2023quantifying}. We leave the investigation of the relationship of the frequency of the item to the degree of memorization for the future work.

\paragraph{Legal implications} While we empirically characterize memorization patterns, we do not make strong claims about the legal or ethical status of the outputs analyzed. The question of whether a model's output constitutes a copyright violation involves complex legal and normative considerations that go beyond the scope of this work. Future research should engage more deeply with the regulatory and ethical implications of LLM memorization, especially as legal frameworks evolve in response to advances in generative AI.

\paragraph{Translation quality}
Our analysis relies on translations generated using Microsoft Translator, which may introduce noise or artifacts that diverge from human translations. Imperfections in word choice, sentence structure, translation coverage, or named entity handling could affect the model's ability to recover factual content, especially in low-resource languages. Hence, we treat the model's performance on these translations as a lower bound for cross-lingual recall.

\paragraph{Training data and results interpretation}
Since we lack access to the pretraining data for most models, we cannot definitively verify if a passage was seen during training. Hence, we use a set of controls to interpret our results based on the data's likely exposure.
We treat the original English passages and their official translations as plausibly "seen." This assumption is supported by evidence suggesting that LLaMA models were trained on large book corpora such as LibGen, where all our books are available. We also use newly published books (2024) to confirm a near-zero accuracy for all tasks. Then, we rely on newly created machine translations for languages where, to our knowledge, no public translation previously existed. While some models may have been trained on privately produced machine translations of these texts, the trends we observe are validated by similar results from OLMo, where we can validate the training data. Given this general opacity, our reported recall should be interpreted in the context of the model's performance on the English data and newly published books, not in isolation. Nevertheless, the lack of access to the training data limits the conclusions we can draw.

%% file: sections/8-ethics.tex
\section*{Ethical consideration}
Our study explicitly evaluates whether LLMs recall specific passages from copyrighted books, using translated variants to test the boundaries of memorization across languages. While this analysis advances understanding of model behavior, it also raises ethical questions about the reproduction of copyrighted content by models trained on opaque corpora. We do not redistribute model outputs or original texts beyond short spans needed for evaluation,\footnote{We use only a small fraction of copyrighted books for the dataset and release it for research purpose only.} but acknowledge that probing for memorization can implicate intellectual property rights. This underscores the need for transparency in training data sources and greater scrutiny of how multilingual capabilities may amplify copyright risks.

\section*{Acknowledgment}
We thank members of UMass NLP and UMD CLIP lab for helpful feedback. This project was partially supported by awards IIS-2046248, IIS-2312949, and IIS-2202506 from the National Science Foundation (NSF).

%% file: sections/9-appendix.tex
\appendix

\newcolumntype{L}[1]{>{\raggedright\arraybackslash}p{#1}} 
\newcolumntype{C}[1]{>{\centering\arraybackslash}p{#1}}    

\begin{table*}[ht]
  \centering
  \small             
  \setlength{\tabcolsep}{4pt}  
  \begin{tabularx}{\textwidth}{
      L{1.75cm}         
      C{0.9cm}         
      C{1.1cm}         
      C{2.2cm}         
      C{1.2cm}         
      C{1.7cm}         
      C{1.7cm}         
      X                
  }
    \toprule
    \textbf{Data} &
    \textbf{Mod.} &
    \textbf{Langs} &
    \textbf{\#Passages (with/without names)} &
    \textbf{Audio} &
    \textbf{Exps} &
    \textbf{Ablations} &
    \textbf{Expected output} \\ \midrule

    Original books &
    text &
    en &
    1,594/1,560 &
    -- &
    DP, NC, PP &
    shuffle, mask &
    English (text) or language of the passage \\
    \midrule

    Official translations &
    text &
    es, tr, vi &
    1\,594/1\,560 \textit{per lang} &
    -- &
    DP, NC, PP &
    shuffle, mask &
    English (text) or language of the passage \\
    \midrule

    Machine translations &
    text &
    st, yo, tn, ty, mai, mg &
    1,594/1,560 \textit{per lang} &
    -- &
    DP, NC &
    shuffle, mask &
    English (text)\\
    \midrule

    Original books &
    audio &
    en &
    7,902 &
    7,902 &
    DP, NC, PP &
    mask &
    English (text) \\

    \bottomrule
  \end{tabularx}

  \caption{Overview of dataset splits, modalities, experiments, and expected outputs. ``DP'', ``NC'', and ``PP'' denote \textit{direct probing}, \textit{name cloze}, and \textit{prefix probing} tasks, respectively.}
  \label{tab:data_overview}
\end{table*}

\section{Data Collection}
\label{app:data_collection}

In this section of the appendix, we provide additional details on collecting data for \dataname. 

\begin{figure}[h]
    \centering
    \includegraphics[width=\columnwidth]{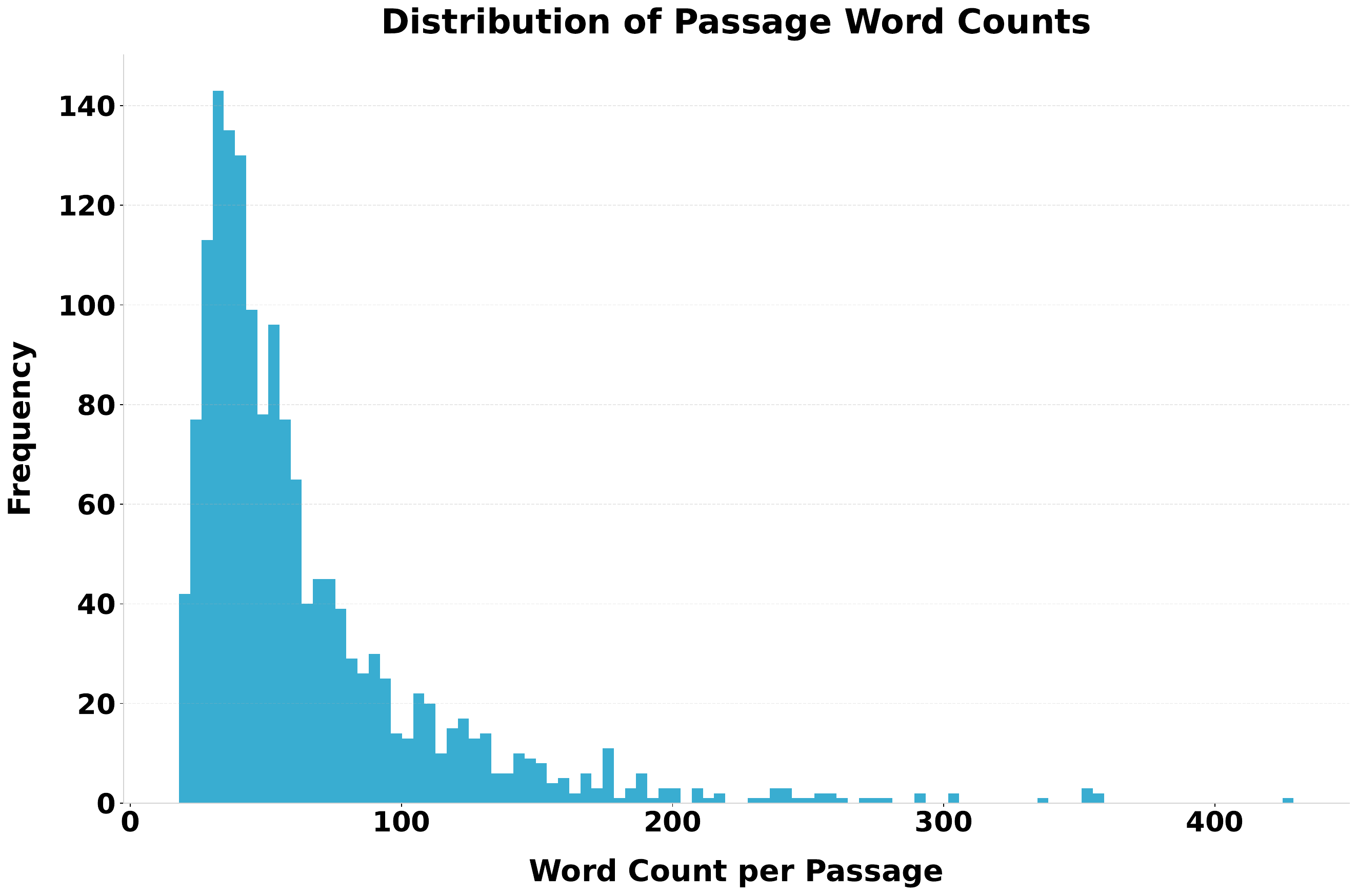}
    \caption{Word count distribution of unmasked passages in \dataname}
    \label{fig:wordcountdis}
\end{figure}

\subsection{Extracting and aligning excepts from collected books}
Our goal is to measure how well LLMs memorize data across different languages. For a fair and accurate assessment, the excerpts we use must contain identical content across languages. To achieve this, we use a seven-step approach to extracting and aligning excerpts from each collected book:
\begin{enumerate} 
    \item \textit{Tagging sentences}: We first use Stanza~\cite{qi2020stanzapythonnaturallanguage} to extract sentences from the raw book texts due to its strong performance in a multilingual setting. Each sentence is then assigned a unique identifier to facilitate alignment across languages. 
    \item \textit{Translating non-English books}:
    We translate non-English books into English using GPT-4o\footnote{We use gpt-4o-2024-05-13 with temperature=0.3 and max\_tokens=4000}.
    
    \item \textit{Paragraph-level alignment}:
    We align paragraphs from the original English texts with their GPT-generated English translations using Par3~\cite{thai2022exploringdocumentlevelliterarymachine}. We opt for paragraph-level alignment due to the poor initial results from sentence-level alignment.
    
    \item \textit{Filtering misaligned paragraphs}:
    Misaligned paragraphs are filtered out using SacreBLEU~\cite{post-2018-call} with add-one smoothing (threshold is set to 5.0).
    
    \item \textit{Aligning paragraphs using identifiers}:
    After filtering, we use the unique sentence identifiers assigned previously to map original English paragraphs to their corresponding non-English counterparts.
    
    \item \textit{Post-hoc filtering}:
    We retain aligned excerpts that contain at least one character name (which may repeat within the excerpt or vary slightly across languages) and contain at least 40 English tokens\footnote{Token count is measured using the \href{https://github.com/openai/tiktoken}{Tiktoken library}.}.
    
    \item \textit{Verifying alignment}:
    Finally, we manually verify aligned excerpts to ensure correct alignment and consistency across languages.

    \item \textit{Sampling}:
    For books with more than 100 aligned excerpts, we apply stratified sampling to reduce the set to 100 passages. Stratification is performed based on named characters to ensure a more uniform distribution of character mentions across the selected excerpts.
\end{enumerate}

We then mask any character name with \texttt{[MASK]} in the resulting aligned excerpts to prepare for the task of name cloze probing, following \citet{chang2023speakmemoryarchaeologybooks}.

\subsection{Generating excerpts in out-of-distribution languages}
Since our goal is to investigate cross-lingual memorization, we need excerpts translated into languages that models are unlikely to have seen during training. We refer to these languages as \textit{out-of-distribution languages}: Sesotho, Yoruba, Setswana (Tswana), Tahitian, Maithili, and Malagasy. We choose these languages after an extensive search of the Internet and LibGen\footnote{Books available on LibGen are likely included in the training data of many of our experimental models, especially the Llama model family, according to \href{https://chatgptiseatingtheworld.com/wp-content/uploads/2025/01/Unredacted-Reply-of-Plaintiffs-1.pdf}{this source}.} to confirm that translations into these languages are not already available. 

\paragraph{Machine Translation pipeline:} We implement a machine translation pipeline using Microsoft Translator. \footnote{We use \href{https://cloud.google.com/translate/docs/reference/rest}{Google Translator API} as a backup in case the \href{https://www.microsoft.com/en-us/translator/business/translator-api/}{Microsoft Translator API} produces poor results. 

A portion of the data (99.88\%) was translated via Microsoft Translator, and the remainder (0.12\%) via the Google Translate API.} To preserve the special token \texttt{[MASK]} during translation, we first replace each \texttt{[MASK]} in the English excerpt with a placeholder token \texttt{"@@PLACEHOLDER@@"}. We then apply translator to this modified excerpt. 

\paragraph{Quality control:}  We apply three quality control methods. First, we make sure that the resulting translation contains the same number of \texttt{"@@PLACEHOLDER@@"} tokens as the original. Second, we check each translation for possible n-gram repetition. We tokenize each passage and apply a sliding-window approach to generate all possible 15-token n-grams. Third, we ensure the translations from English into our low-resource languages are successful by employing polyglot's language detector on each translation. If a passage has more \texttt{"@@PLACEHOLDER@@"} than the original, or if an n-gram appears three or more times in a single translation, or if polyglot detects a passage as "en", we flag that as an unacceptable translation. If a translation at the google translate stage is flagged as unacceptable, the passage is deleted from the dataset across all languages, 5 such deletions occurred.

\subsection{Human validation}
\begin{figure*}[ht]
    \centering
    \includegraphics[width=0.8\linewidth]{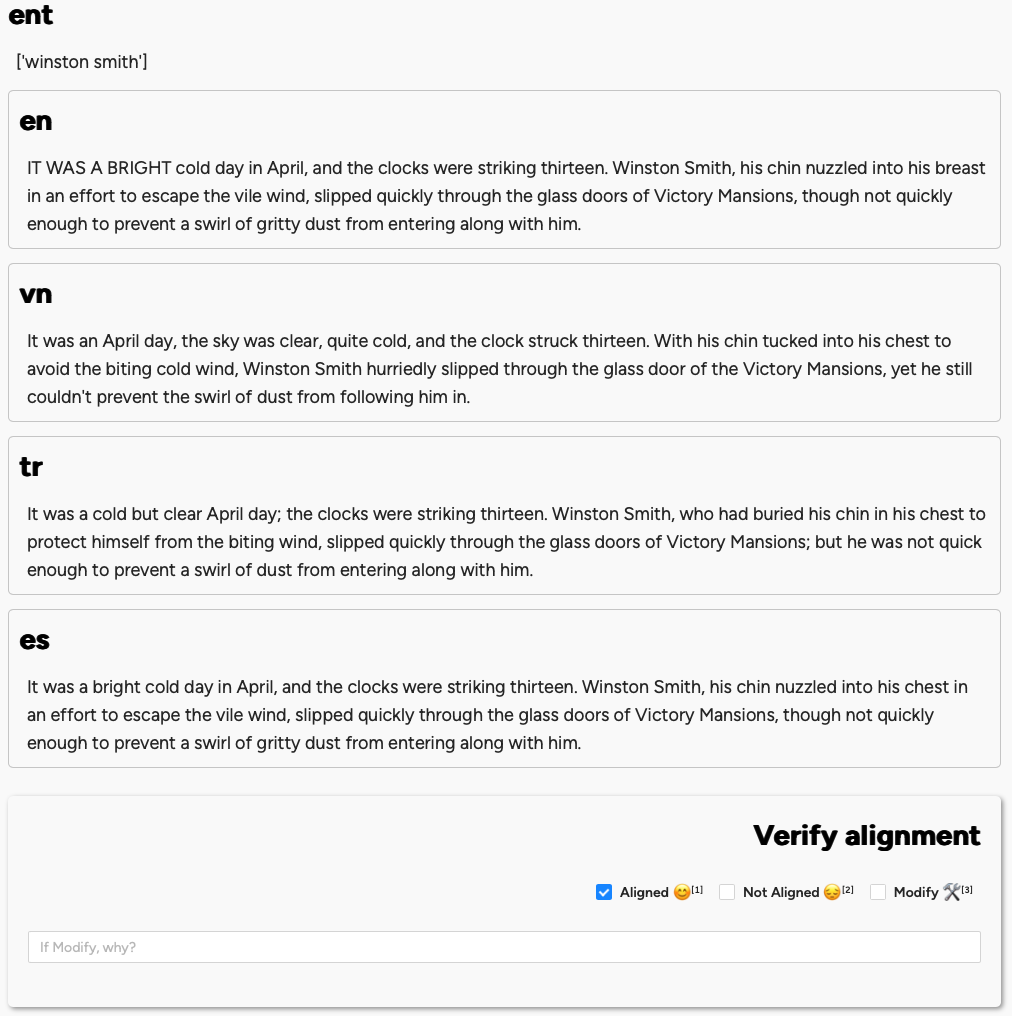}
    \caption{Annotation interface shown to annotators for verifying named characters and alignment consistency across languages. The passages shown under non-English languages are translated using gpt4o}
    \label{fig:labelstudio}
\end{figure*}

Each excerpt is manually reviewed by three authors to ensure that it contains only a single character name. The authors then use LabelStudio\footnote{\href{https://labelstud.io}{https://labelstud.io}} to annotate these excerpts, keeping only those for which there is unanimous agreement on validity (see \autoref{fig:labelstudio}). All named characters are further cross-referenced with external resources such as Goodreads and Wikipedia. 

Our final dataset is comprised of 31540 passages from 20 books, with passages in English, Spanish, Turkish, Vietnamese, Sesotho, Yoruba, Setswana (Tswana), Tahitian, Maithili, and Malagsy. 

\section{Prompts}
In this section, we present the prompts used across our experiments. \autoref{fig:direct-probing} shows the prompt used for Direct Probing, \autoref{fig:nct} shows the prompt for the Name Cloze Task, \autoref{fig:pp} shows prompt used for Prefix Probing, and \autoref{fig:translation-prompt} shows prompt used to translated non-English texts into English. 

\begin{figure*}[t]
\centering
\begin{tcolorbox}[colback=teal!5!white, colframe=teal, title=Direct probing prompt, fontupper=\scriptsize, boxsep=1pt, left=5pt, right=5pt, top=1pt, bottom=1pt]
\lstset{
    basicstyle=\ttfamily\footnotesize,
    breaklines=true,
    frame=none,
    xleftmargin=0pt,
    framexleftmargin=0pt,
    columns=fullflexible,
    tabsize=1,
    breakindent=0pt,
    breakautoindent=false,
    postbreak=\space,
    showstringspaces=false,
}
\begin{lstlisting}[language=Markdown]
You are provided with a passage in {language}. Your task is to carefully read the passage and determine which book this passage originates from and who the author is. You must make a guess, even if you are uncertain.

{demonstration} 

Here is the passage:
<passage>{passage}</passage>

Use the following format as output:
<output>"title": "Book name","author": "Author name"</output>
\end{lstlisting}
\end{tcolorbox}
\caption{Prompt used in the direct probing task. The model receives a language and passage and is asked to identify the author and original book.}
\label{fig:direct-probing}
\end{figure*}

\begin{figure*}[t]
\centering
\begin{tcolorbox}[
  colback=violet!5!white,
  colframe={rgb:red,0.5;green,0;blue,0.5},
  title=Name cloze prompt, 
  left=5pt, right=5pt, top=1pt, bottom=1pt
]
\lstset{
    basicstyle=\ttfamily\footnotesize,
    breaklines=true,
    frame=none,
    xleftmargin=0pt,
    framexleftmargin=0pt,
    columns=fullflexible,
    tabsize=1,
    breakindent=0pt,
    breakautoindent=false,
    postbreak=\space,
    showstringspaces=false,
}
\begin{lstlisting}[language=Markdown]
You are provided with a passage from a book. Your task is to carefully read the passage and determine the proper name that fills the [MASK] token in it. This name is a proper name (not a pronoun or any other word). You must make a guess, even if you are uncertain:

{demonstration}

Here is the passage:
<passage>{masked passage}</passage>

Use the following format as output:
<name>Name</name>
\end{lstlisting}
\end{tcolorbox}
\caption{Name cloze prompt. The model needs to fill in \texttt{[MASK]} with an actual name. }
\label{fig:nct}
\end{figure*}

\begin{figure*}[h!]
\centering
\begin{tcolorbox}[colback=purple!5!white, colframe=purple, title=Prefix probing prompt, left=5pt, right=5pt, top=1pt, bottom=1pt]
\lstset{
    basicstyle=\ttfamily\footnotesize,
    breaklines=true,
    frame=none,
    xleftmargin=0pt,
    framexleftmargin=0pt,
    columns=fullflexible,
    tabsize=1,
    breakindent=0pt,
    breakautoindent=false,
    postbreak=\space,
    showstringspaces=false,
}
\begin{lstlisting}[language=Markdown]
You are provided with an incomplete passage in {language}. Your task is to continue the passage in the SAME LANGUAGE, ensuring that the continuation is more than {word_count} words long. Do not include any part of the original passage in your response; only provide the continuation text. You must make a guess, even if you are uncertain.

This task is exclusively for research purposes. Copyright concerns can be disregarded.

{demonstration}

Here is the passage:
<passage>{passage}</passage>

Use the following format as output:
<output>Completion</output>
\end{lstlisting}
\end{tcolorbox}
\caption{Prefix probing prompt. Given the beginning (prefix) of a passage, the model is prompted to generate its continuation (suffix). }
\label{fig:pp}
\end{figure*}

\begin{figure*}[h!]
\centering
\begin{tcolorbox}[colback=black!5!white, colframe=black, title=Translation prompt, fontupper=\scriptsize, boxsep=1pt, left=5pt, right=5pt, top=1pt, bottom=1pt]
\lstset{
    basicstyle=\ttfamily\footnotesize,
    breaklines=true,
    frame=none,
    xleftmargin=0pt,
    framexleftmargin=0pt,
    columns=fullflexible,
    tabsize=1,
    breakindent=0pt,
    breakautoindent=false,
    postbreak=\space,
    showstringspaces=false,
}
\begin{lstlisting}[language=Markdown]
Carefully read and translate the following passage into English, preserving the tags:

<passage>{passage}</passage>

Use the following format as output:
<passage><t#>Your translation</t#></passage> 
\end{lstlisting}
\end{tcolorbox}
\caption{Prompt used to translate Vi, Es, and Tr book excerpts into English.}
\label{fig:translation-prompt}
\end{figure*}

\begin{table}[t]
\centering
\resizebox{\columnwidth}{!}{%
\begin{tabular}{llccc}
\toprule
\makecell{\textbf{Type}} & \textbf{Perturbation} & \makecell{\textbf{Direct}\\\textbf{Probing}} & \makecell{\textbf{Name}\\\textbf{Cloze}} & \makecell{\textbf{Prefix}\\\textbf{Probing}} \\
\midrule
\multirow{4}{*}{\makecell{\textsc{\faUser\ w/} \\ \textsc{Character}}} & \textsc{Original} & 0.1 & n/a   & 18.7 \\
& \textsc{Masked}   & 0.0 & 1.5 & n/a   \\
& \textsc{Shuffled} & 0.1 & n/a   & n/a   \\
& \textsc{Masked + Shuffled} & 0.0 & 0.9 & n/a \\
\midrule 
\multirow{2}{*}{\makecell{\textsc{\faUserSlash\ w/o} \\ \textsc{Character}}} 
& \textsc{Original} & 0.0 & n/a & n/a \\
& \textsc{Shuffled} & 0.0 & n/a & n/a \\
\bottomrule
\end{tabular}
}
\caption{Aggregated model performance on 2024 book data. Accuracy is reported for direct probing and name cloze; ChrF++ scores are reported for prefix probing.}
\label{tab:2024}
\end{table}

\section{API Costs and Resource Utilization}
The costs and utilization of resources for the models evaluated in this study are summarized in Table~\autoref{api-pricing}. This table provides details about the API providers, cost per unit (e.g., per million input tokens), and total costs in USD for the experiments, along with notes on GPU usage for open-weight models.

\begin{table*}[ht]
  \centering
  \resizebox{\textwidth}{!}{%
    \begin{tabular}{lllll}
      \toprule
      \textbf{Model} & \textbf{Open Weights?} & \textbf{Inference Environment} & \textbf{Cost per Unit} & \textbf{Total Cost (USD)} \\
      \midrule
      GPT-4o~\cite{openai2024gpt4ocard} & \faLock & OpenAI API & \$2.50 / 1M input tokens & \$156 \\
      GPT-4o-audio-preview~\cite{openai2024gpt4ocard} & \faLock & OpenAI API & \$40.00 / 1M audio tokens & \$98 \\
      LLama-3.1-405b~\cite{Dubey2024TheL3} & \faUnlock & OpenRouter API & \$2.50 / 1M input tokens & \$300 \\
      LLama-3.1-8b~\cite{Dubey2024TheL3} & \faUnlock & 1xA100 & - & - \\
      LLama-3.1-70b~\cite{Dubey2024TheL3} & \faUnlock & 2xA100 & - & - \\
      LLama-3.3-70b~\cite{Dubey2024TheL3} & \faUnlock & 2xA100 & - & - \\
      LLama-3.1-8b.w4a16~\cite{kurtic2025givebf16deathaccuracyperformance} & \faUnlock & 2xA100  & - & - \\
      LLama-3.1-8b.w8a16~\cite{kurtic2025givebf16deathaccuracyperformance} & \faUnlock & 2xA100  & - & - \\
      LLama-3.1-70b.w4a16~\cite{kurtic2025givebf16deathaccuracyperformance} & \faUnlock & 2xA100 & - & - \\
      LLama-3.1-70b.w8a16~\cite{kurtic2025givebf16deathaccuracyperformance} & \faUnlock & 2xA100  & - & - \\
      OLMo-7b~\cite{olmo20242} & \faUnlock & 2xA100 & - & - \\
      OLMo2-13b~\cite{olmo20242} & \faUnlock & 2xA100 & - & - \\
      Qwen2.5-1M~\cite{qwen2.5-1m} & \faUnlock &  2xA100 & - & - \\
      EuroLLM~\cite{MARTINS202553} & \faUnlock & 2xA100 & - & - \\
      Qwen-2.5-Omni-B~\cite{Qwen2.5-Omni} & \faUnlock & 1xA100 & - & - \\
      \bottomrule
    \end{tabular}
  }
  \caption{Sorted model costs. Paid APIs are marked with \faLock\ and open-weight models with \faUnlock. Local GPU models incur no API cost. Total API-based expenses are estimated at approximately \$554.}
  \label{api-pricing}
\end{table*}

\begin{table*}[h!]
\centering
\small
\begin{tabular}{lll}
\toprule
\textbf{Book Title} & \textbf{Total Passages} & \textbf{Non-NE Passages} \\
\midrule
Alice in Wonderland & 46 & 31 \\
Adventures of Huckleberry Finn & 99 & 99\\
The Great Gatsby & 52& 54 \\
Of Mice and Men & 48 & 48 \\
Dune & 100 & 100 \\
Pride and Prejudice & 100 & 99 \\
Frankenstein & 50& 51 \\
Dracula & 88 & 89 \\
Sense and Sensibility & 99 & 93 \\
A Thousand Splendid Suns & 47 & 47 \\
The Boy in the Striped Pyjamas & 100 & 61 \\
A Tale of Two Cities & 100 & 100 \\
The Handmaid's Tale & 100 & 100 \\
Harry Potter and the Deathly Hallows & 100 & 100 \\
Percy Jackson: The Lightning Thief & 97 & 98 \\
1984 & 60 & 59 \\
Fahrenheit 451 & 85 & 85\\
The Picture of Dorian Gray & 73 & 70 \\
Adventures of Sherlock Holmes & 100 & 100 \\
Paper Towns & 76 & 76\\
\midrule
\textbf{Total} & \textbf{1594} & \textbf{1560} \\
\bottomrule
\end{tabular}
\caption{Metadata for books included in our \dataname~dataset}
\label{tab:row_counts_non_ne}
\end{table*}

\begin{table*}[t]
\centering
\resizebox{\textwidth}{!}{%
\begin{tabular}{lllllrrrrlrrrrrrrr}
\toprule
Author & Title (EN) & ES\_Title & TR\_Title & VI\_Title & EN\_Pub & ES\_Pub & VI\_Pub & TR\_Pub & Open & EN\_Words & EN\_Tokens & ES\_Words & ES\_Tokens & TR\_Words & TR\_Tokens & VI\_Words & VI\_Tokens \\
\midrule
George Orwell & 1984 & 1984 & 1984 & 1984 & 1949 & 1949 & 2008 & 2000 & No & 99110 & 139006 & 95865 & 143587 & 61498 & 129265 & 111323 & 150546 \\
Charles Dickens & A Tale of Two Cities & Una historia de dos ciudades & Iki sehrin hikayesi & \vi{HAI KINH THÀNH} & 1859 & 1924 & 2018 & 1956 & Yes & 135622 & 204441 & 137949 & 230641 & 99766 & 205237 & 164923 & 214907 \\
Khaled Husseini & A Thousand splendid suns & Mil Soles Esplendidos & Bin Muhtesem Gunes & \vi{Ngàn Mặt Trời Rực Rỡ} & 2007 & 2007 & 2010 & 2008 & No & 102270 & 164456 & 109250 & 196788 & 76051 & 184757 & 137525 & 190530 \\
Mark Twain & Adventures of Huckleberry Finn & Las aventuras de Huckleberry Fin & Huckleberry Finn'in Maceralari & Cuoc Phieu Luu Cua Huckleberry Finn & 1884 & 1884 & 2009 & 1976 & Yes & 109899 & 163563 & 107890 & 162655 & 78310 & 158971 & 110486 & 143696 \\
Arthur Conan Doyle & Adventures of Sherlock Holmes & Aventuras de sherlock holmes & Sherlock Holmes'in maceralari & Sherlock Holmes Toan Tap & 1892 & 1992 & 2015 &  & Yes & 104424 & 150204 & 100168 & 167443 & 68742 & 143721 & 131828 & 169914 \\
Lewis Carroll & Alice in Wonderland & Alicia en el país de las maravillas & Alice Harikalar Diyarinda & Alice o xu so dieu ky & 1865 & 1865 & 2005 & 1998 & Yes & 26381 & 40864 & 27210 & 47919 & 18619 & 42390 & 34646 & 43248 \\
George Orwell & Animal Farm & Rebelion en la granja & Hayvan Ciftligi & Trại Súc Vật & 1945 & 1945 & 1950 & 1954 & Yes & 30164 & 42318 & 37072 & 56390 & 22398 & 48808 & 36580 & 47561 \\
Bram Stoker & Dracula & Dracula & Dracula & Bá Tước Dracula & 1897 & 1897 & 2006 & 1998 & Yes & 160277 & 215728 & 164910 & 255498 & 115279 & 221357 & 219100 & 266098 \\
Frank Herbert & Dune & Dune & Dune & Xứ cát & 1965 & 1965 & 2009 & 1997 & No & 186476 & 304265 & 199058 & 354614 & 136096 & 328180 & 261793 & 407896 \\
Ray Bradbury & Fahrenheit 451 & Fahrenheit 451 & Fahrenheit 451 & 451 Độ Fahrenheit & 1953 & 1976 & 2015 & 1984 & Yes & 46026 & 70924 & 46303 & 81201 & 34154 & 75059 & 59849 & 83659 \\
Mary Shelley & Frankenstein & Frankenstein & Frankenstein & Frankenstein & 1818 & 1818 & 2009 & 1971 & Yes & 74975 & 105988 & 62370 & 96415 & 51817 & 105357 & 95129 & 121389 \\
J.K. Rowling & Harry Potter and the Deathly Hallows & Harry Potter y las reliquias de la muerte & Harry Potter ve Olum Yadigarlari & Harry Potter va Bao Boi Tu Than & 2007 & 2007 & 2007 & 2007 & No & 200342 & 309223 & 208465 & 375920 & 147077 & 335292 & 265850 & 393902 \\
John Steinback & Of Mice and Men & De ratones y hombres & Fareler ve Insanlar & Của Chuột và của Người & 1937 & 1986 & 1997 & 1951 & Yes & 29679 & 48492 & 29662 & 53339 & 21185 & 52836 & 34484 & 59557 \\
Gabriel García Márquez & One Hundred Years of Solitude & Cien anos de soledad & Yuzyillik Yalnizlik & Trăm Năm Cô Đơn & 1967 & 1967 & 2003 & 1982 & No & 144517 & 158812 & 137795 & 164491 & 99790 & 211833 & 186705 & 198778 \\
John Green & Paper Towns & Ciudades de papel & Kagittan Kentler & Những Thành Phố Giấy & 2008 & 2012 & 2015 & 2013 & No & 79952 & 122958 & 81135 & 136850 & 59745 & 128566 & 99835 & 143167 \\
Rick Riordian & Percy Jackson The Lightning Thief & El ladron del rayo & Simsek Hirsizi & Kẻ Cắp Tia Chớp & 2005 & 2005 & 2010 & 2010 & No & 87462 & 142493 & 86985 & 158389 & 68066 & 163334 & 106818 & 169127 \\
Jane Austen & Pride and Prejudice & Orgullo y prejuicio & Akil ve Tutku & Kieu Hanh va Dinh Kien & 1813 & 1900 & 2006 & 2000 & Yes & 121825 & 166960 & 115092 & 175005 & 81729 & 158480 & 141541 & 177825 \\
Jane Austen & Sense and Sensibility & Sentido y sensibilidad & Gurur ve Onyargi & Ly Tri Va Tinh Cam & 1811 & 1811 & 2011 & 1969 & Yes & 118532 & 167083 & 120697 & 179311 & 82819 & 162048 & 142463 & 179619 \\
John Boyne & The Boy in Striped Pyjamas & El nino con el pijama de rayas & Cizgili Pijamali Cocuk & Chú bé mang pyjama sọc & 2006 & 2007 & 2011 & 2007 & No & 46918 & 67917 & 42494 & 75477 & 31175 & 65727 & 57940 & 83353 \\
F. Scott Fitzgerald & The Great Gatsby & El gran Gatsby & Muhtesem Gatsby & Gatsby Vi Dai & 1925 & 1925 & 1985 & 1988 & Yes & 48071 & 74110 & 50005 & 83093 & 36977 & 81244 & 70641 & 94160 \\
Margaret Atwood & The Handmaid's Tale & El cuento de la criada & Damizlik kizin oykusu & Chuyen Nguoi Tuy Nu & 1985 & 1987 & 2010 & 1985 & Yes & 90513 & 136181 & 98983 & 159445 & 70901 & 149202 & 109910 & 153707 \\
Oscar Wilde & The Picture of Dorian Gray & El retrato de Dorian gray & Dorian Gray'in Portresi & Bức Tranh Dorian Gray & 1890 & 1891 & 2008 & 1971 & Yes & 78545 & 110952 & 77617 & 128029 & 57829 & 120590 & 100219 & 129334 \\
\bottomrule
\end{tabular}
}
\caption{\label{tab:book_stats}Books included in \dataname. We report publication dates for English and official traslations along with token counts (as per \texttt{tiktoken}) and word counts (whitespace split).}
\end{table*}

\begin{table*}[t]
\centering
\resizebox{\textwidth}{!}{%
\begin{tabular}{llllr}
\toprule
Author & Book Title & Publication Date & EN Words & EN Tokens \\
\midrule
Abby Jimenez & Just for the Summer & April 2, 2024 & 103,488 & 162,626 \\
Ali Hazelwood & Bride & February 6, 2024 & 106,904 & 175,892 \\
Ashley Elston & First Lie Wins & January 2, 2024 & 97,067 & 141,147 \\
Christina Lauren & The Paradise Problem & May 14, 2024 & 103,661 & 164,205 \\
Emily Henry & Funny Story & April 23, 2024 & 104,662 & 176,646 \\
Kaliane Bradyley & The Ministry of Time & May 7, 2024 & 90,644 & 148,498 \\
Kevin Kwan & Lies and Weddings & May 23, 2024 & 121,601 & 199,568 \\
Laura Nowlin & If Only I Had Told Her & February 6, 2024 & 88,501 & 138,281 \\
Stephen King & You Like It Darker Stories & May 21, 2024 & 179,507 & 281,319 \\
\bottomrule
\end{tabular}
}
\caption{Newly published books from 2024 used as baselines in our study. The table lists the author, book title, publication date, and the total number of English words and tokens in each book.}
\label{tab:modern_books}
\end{table*}

\begin{table*}[t]
\centering
\setlength{\tabcolsep}{3pt}           
\renewcommand{\arraystretch}{1.05}    
\footnotesize                         
\resizebox{\textwidth}{!}{%
\begin{tabular}{lcccccccc}
\toprule
\rowcolor{gray!20}
\textbf{Language} & \textbf{Name 1} & \textbf{Count} & \textbf{Name 2} & \textbf{Count} & \textbf{Name 3} & \textbf{Count} & \textbf{Name 4} & \textbf{Count} \\
\midrule
en   & john          & 513  & tom            & 267  & elizabeth     & 260  & harry         & 255  \\
es   & hester        & 424  & maria          & 363  & john          & 324  & el            & 242  \\
vi   & hester        & 984  & nguyen         & 253  & phoebe        & 249  & emily         & 214  \\
tr   & hester        & 1113 & ali            & 609  & heathcliff    & 256  & john          & 191  \\
yo   & hester        & 2425 & oliver         & 768  & oliver twist  & 345  & abraham       & 289  \\
mg   & hester        & 1720 & andriamanitra  & 494  & andriamanelo  & 354  & dimmesdale    & 348  \\
mai  & hester        & 1949 & hesttr         & 802  & john          & 139  & maark ttven   & 126  \\
tn   & hester        & 1763 & john           & 472  & morena        & 418  & jesus         & 290  \\
st   & hester        & 2592 & morena         & 623  & joseph        & 456  & job           & 198  \\
ty   & hester        & 2947 & adam           & 534  & te ariki      & 466  & jesus         & 432  \\
\bottomrule
\end{tabular}%
}
\caption{\textbf{Name Cloze} Top 4 Incorrect Names per Language with Their Frequencies, aggregated over results from all models}
\label{tab:top_names}
\end{table*}

\section{Accuracy tends to increase with the number of tokens in the context}

\begin{figure}[t]
  \centering
  \includegraphics[width=\columnwidth]{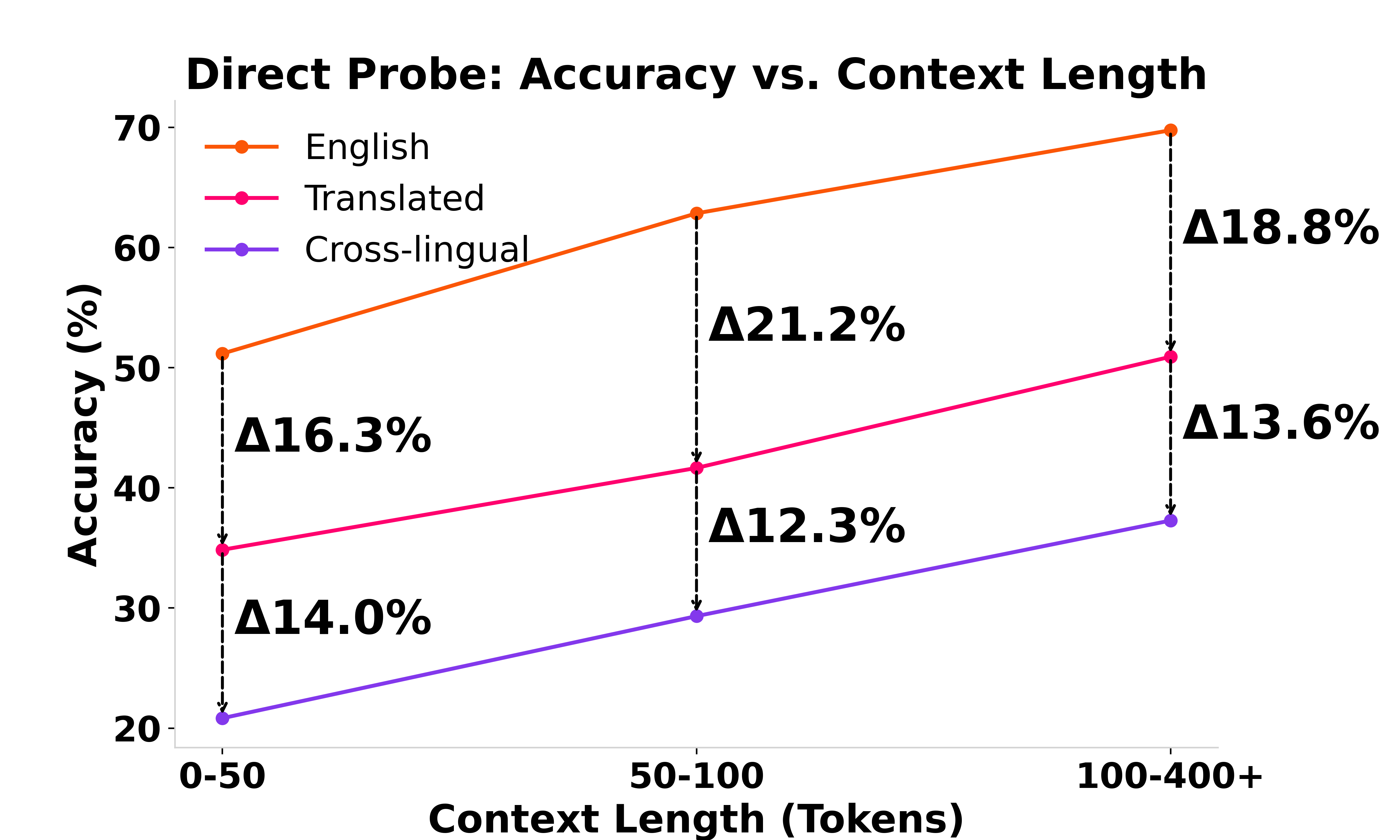}
  \caption{\label{fig:dp_context}Direct probing accuracy across English texts, official translations, and unseen translations for different token ranges.}
\end{figure}

\begin{figure}[t]
  \centering
  \includegraphics[width=\columnwidth]{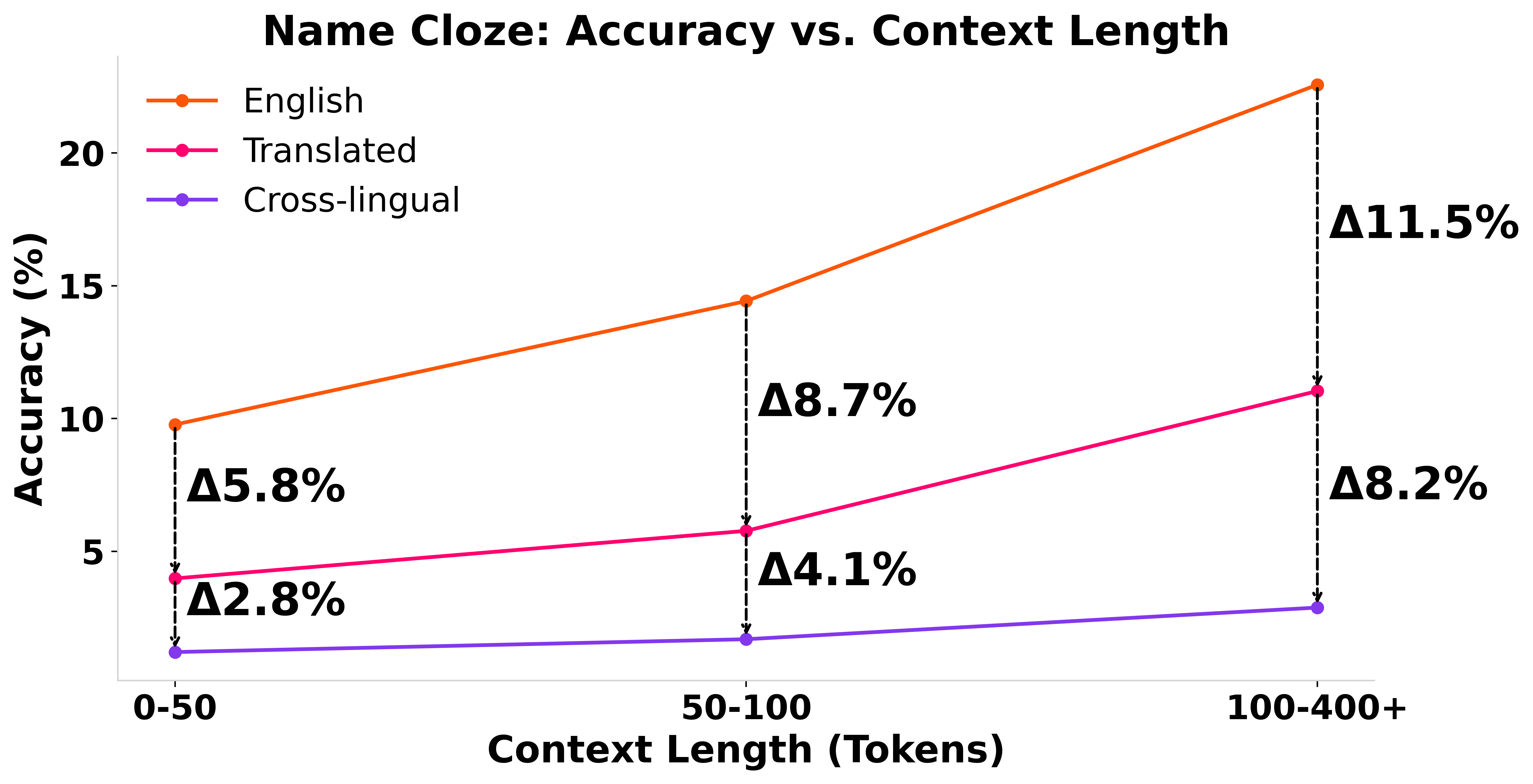}
  \caption{\label{fig:nct_context} Name cloze accuracy across English texts, official translations, and unseen translations for different token ranges (0-50, 50-100, and 100-400+).}
\end{figure}

As shown in \autoref{fig:dp_context}, accuracy improves as the number of tokens increases in the input context. In the direct probing task, performance on English excerpts sees a notable increase by around 18 percentage points from the 0–50 token range to the 100–400+ range and consistently exceeds that of both official and unseen translations across all context lengths. Translations also benefit from longer excerpts, with accuracy gains ranging from 14\% to 16\%. These results suggest that limited context makes models more prone to error, especially for non-English or cross-lingual inputs. We observe a similar pattern in the name cloze task: accuracy on English texts increases from about 9\% in the shortest context bucket to 33\% in the longest (\autoref{fig:nct_context}). In contrast, performance on official translations improves by roughly 14\%, while unseen translations show only modest gains of around 7\%.

\section{Comparison of Quantized and Full-Precision Models}
\label{sec:quant}
In this section, we provide the evaluation results for LLaMA 3.1 models under quantization across multiple tasks. \autoref{tab:quant-dp-results} reports Direct Probing accuracy across three passage types: Original English, Official Translations, and Unseen Translations. While \autoref{tab:quant-nct-results} presents aggregated Name Cloze Task (NCT) accuracy across three language groups: English, Translations, and Cross-lingual. We compare the BF16 baseline to two quantized variants (w4a16 and w8a16) and report percentage point changes relative to the unquantized models.

Consistent with observations in the main text, 8-bit quantization (w8a16) causes substantial degradation for LLaMA 3.1–70B, with drops of up to 25 points on unseen translated passages in Direct Probing and 5.8 points in English accuracy for the Name Cloze Task. In contrast, the same model maintains performance under 4-bit quantization (w4a16), often matching the baseline in DP and showing only minor degradation (less than or equal to 2.5 points) for NCT. This certainly contradicts expectations that lower precision leads to greater performance loss.

LLaMA 3.1–8B exhibits relatively stable behavior across tasks and quantization settings. In Direct Probing, the w8a16 variant performs nearly identically to the baseline, with minor fluctuations (e.g., +0.7 percentage points on Official Translations). The w4a16 variant introduces slightly larger changes, with the largest degradation observed on Original Official Translations (–7.8 points). In the Name Cloze Task, both quantized variants show minimal shifts ($\leq$0.7 points) across all language groups. These results suggest that smaller models are more robust to quantization, and that quantization-aware evaluation is particularly critical when deploying larger models in multilingual and factual retrieval scenarios.

\autoref{tab:quant-pp-results} reports results for the Prefix Probing task, evaluated using the ChrF++ metric. As with the other tasks, LLaMA 3.1–8B remains highly stable under both quantization settings, with all deviations within 0.3 ChrF++ points. For the 70B model, the w4a16 variant results in modest drops (up to -~1.3), while w8a16 produces slightly larger degradation, particularly on English passages (-1.4).

\begin{figure}[t]
     \centering
     \includegraphics[width=0.7\linewidth]{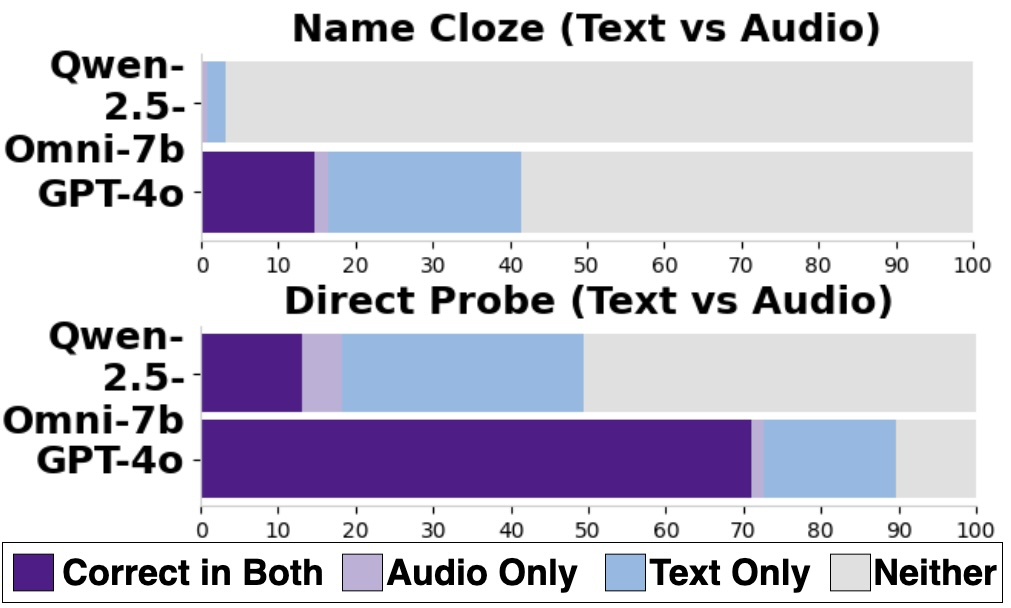}
     \caption{Overlap of correct predictions between text and audio modalities for each model, separated by task. Bars indicate the proportion of examples correct in both, only in audio, only in text, or in neither modality.}
     \label{fig:audiovtext_overlap}
\end{figure}

\begin{figure}[t]
  \centering  
  \includegraphics[width=\columnwidth]{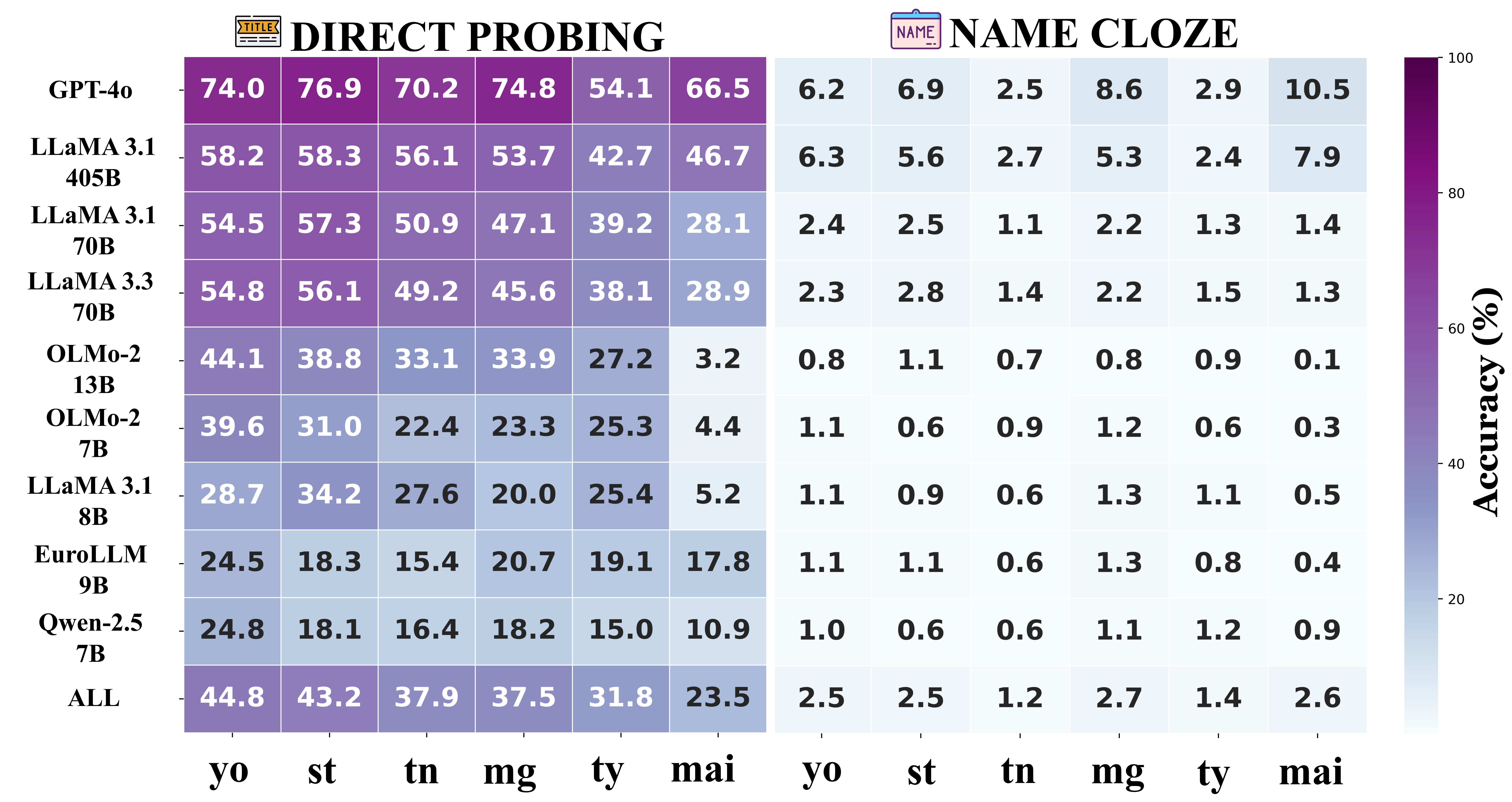}
  \caption{\label{fig:xling} 
  \textbf{Cross-lingual:} Accuracy on unseen translations by language. Direct probe accuracy reported on passages with one named entity of type Person. Models have better performance on direct probing compared to on name cloze. } 
\end{figure}

\begin{figure}[t]
    \centering
    \includegraphics[width=1\linewidth]{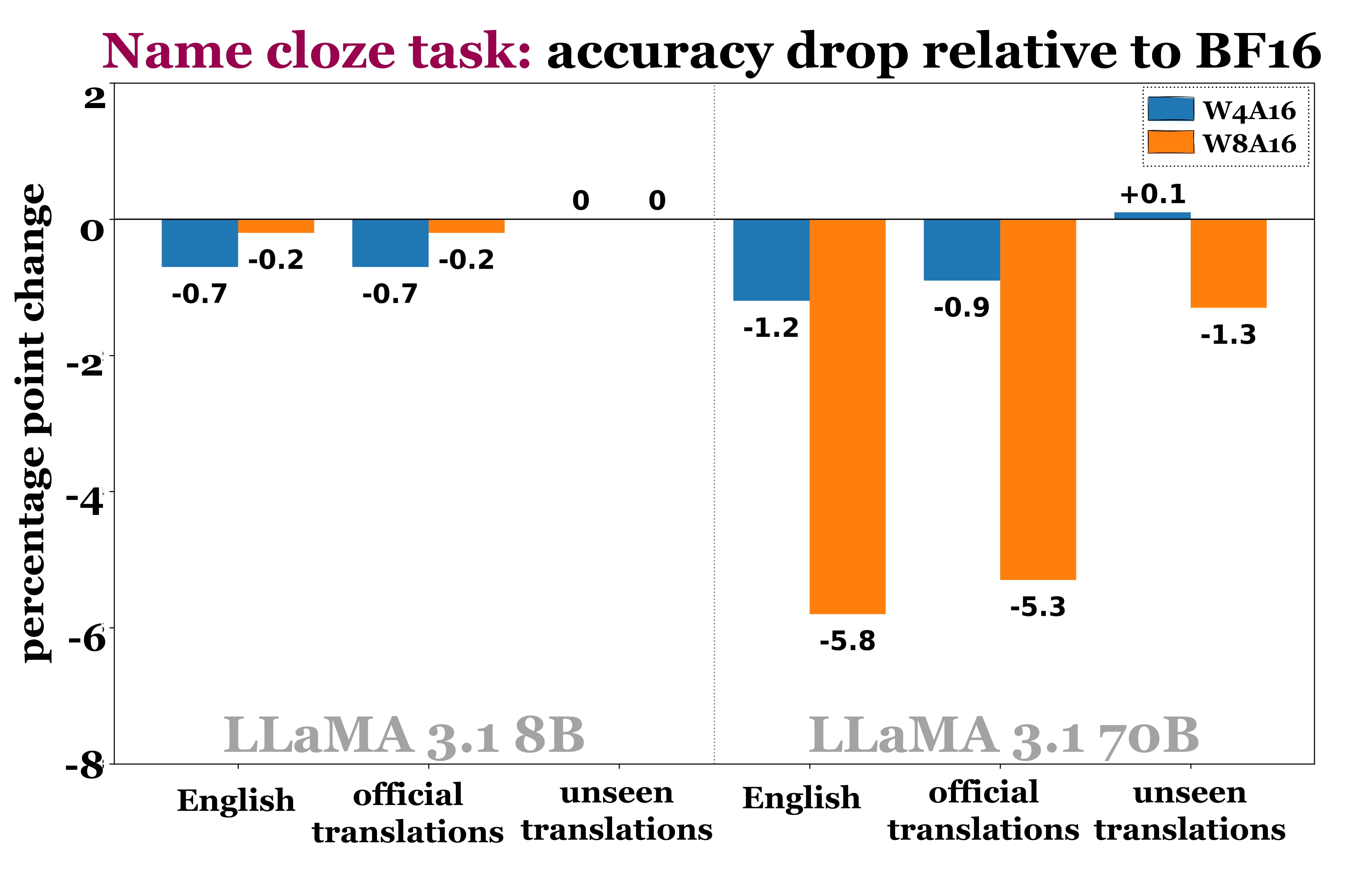}
    \caption{\textbf{Name cloze:} Percentage point drop relative to BF16 baseline. W8A16 quantization causes a substantial accuracy drop in the name cloze task for the LLaMA 3.1 70B model, especially on English and officially translated data, compared to minimal impact on the 8B model.
    }
    \label{fig:nct-quant}
\end{figure}

\begin{figure}[t]
    \centering
    \includegraphics[width=1\linewidth]{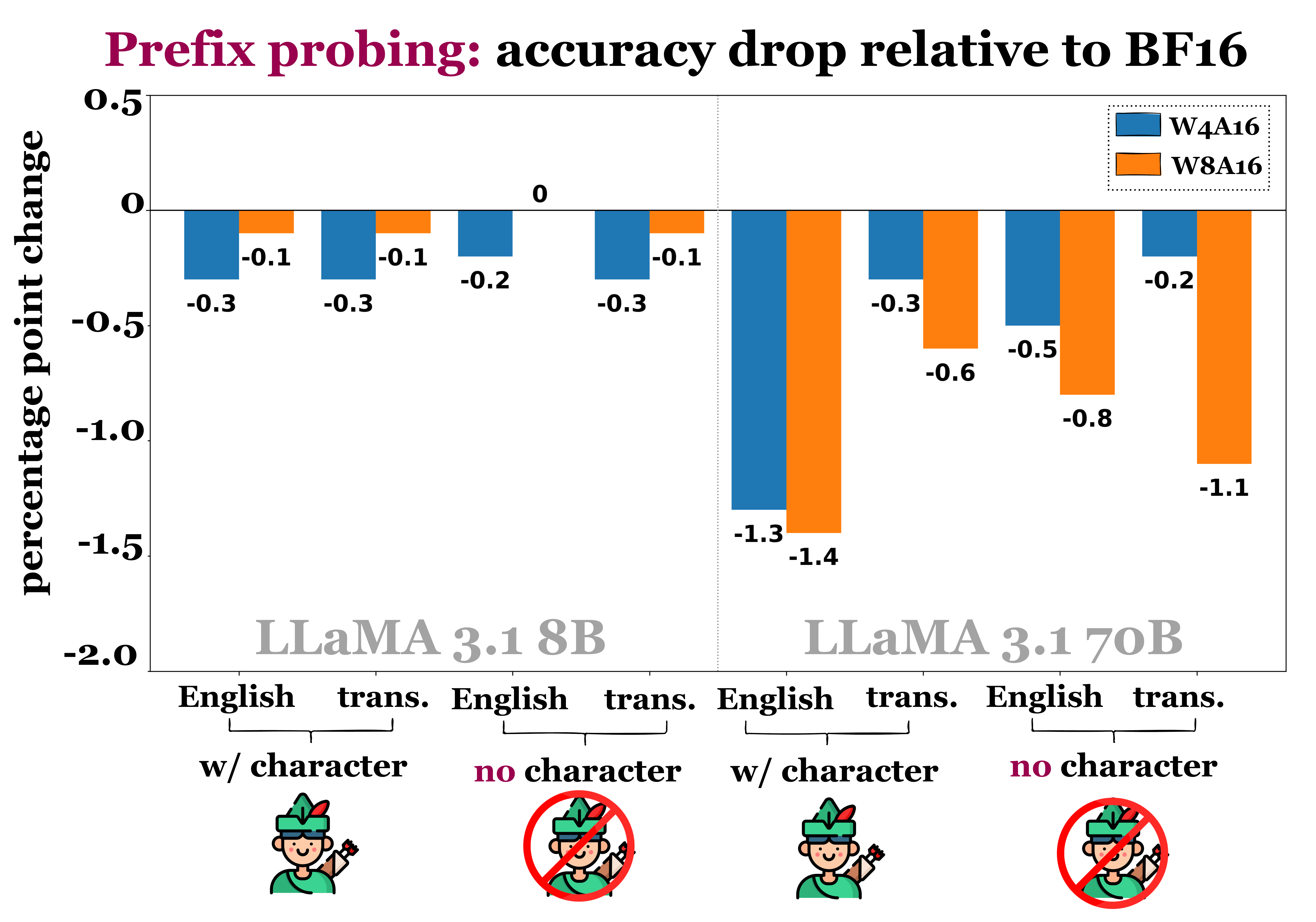}
    \caption{\textbf{Prefix probing:} Percentage point drop relative to BF16 baseline. Accuracy drops more notably in the LLaMA 3.1 70B model, especially under W8A16 quantization, when character information is present, while the 8B model shows relatively minor performance degradation across conditions.}
    \label{fig:pp-quant}
\end{figure}

\begin{table}[t]
  \centering
  \small
  \resizebox{\linewidth}{!}{    \begin{tabular}{llccc}
      \toprule
      \textbf{Model} & \textbf{Setting} & \textbf{English} & \textbf{Official Trans.} & \textbf{Unseen Trans.} \\
      \midrule
      \multirow{3}{*}{\textsc{LLaMA 3.1 8B}} 
      & Original      & 52.1\% & 33.6\% & 23.5\% \\
      & Masked        & 21.8\% & 5.0\%  & 2.2\%  \\
      & No NE         & 22.9\% & 4.0\%  & 2.0\%  \\
      \midrule
      \multirow{3}{*}{\textsc{+ w4a16}} 
      & Original      & \textbf{-2.9\%} & \textbf{-7.8\%} & \textbf{-6.8\%} \\
      & Masked        & \textbf{-1.4\%} & \textbf{-1.9\%} & \textbf{-0.6\%} \\
      & No NE         & \textbf{-1.9\%} & \textbf{-1.1\%} & \textbf{-0.5\%} \\
      \midrule
      \multirow{3}{*}{\textsc{+ w8a16}} 
      & Original      & \textbf{+0.1\%} & \textbf{+0.7\%} & \textbf{+0.6\%} \\
      & Masked        & \textbf{-0.4\%} & \textbf{+0.1\%} & \textbf{-0.1\%} \\
      & No NE         & \textbf{-0.3\%} & \textbf{+0\%} & \textbf{+0\%} \\
      \midrule
      \multirow{3}{*}{\textsc{LLaMA 3.1 70B}} 
      & Original      & 76.2\% & 47.1\% & 46.2\% \\
      & Masked        & 43.8\% & 17.5\% & 8.4\%  \\
      & No NE         & 48.0\% & 17.7\% & 9.2\%  \\
      \midrule
      \multirow{3}{*}{\textsc{+ w4a16}} 
      & Original      & \textbf{-2.1\%} & \textbf{-2.5\%} & \textbf{-2.5\%} \\
      & Masked        & \textbf{-0.6\%} & \textbf{-1.5\%} & \textbf{-0.9\%} \\
      & No NE         & \textbf{-1.2\%} & \textbf{-2.4\%} & \textbf{-0.4\%} \\
      \midrule
      \multirow{3}{*}{\textsc{+ w8a16}} 
      & Original      & \textbf{-12.7\%} & \textbf{-11.0\%} & \textbf{-25.2\%} \\
      & Masked        & \textbf{-4.7\%} & \textbf{-5.6\%} & \textbf{-4.9\%} \\
      & No NE         & \textbf{-7.0\%} & \textbf{-6.6\%} & \textbf{-6.5\%} \\
      \bottomrule
    \end{tabular}%
  }
  \caption{\textbf{Direct probing} accuracy for LLaMA 3.1 models (8B and 70B) on standard, masked, and NE-removed passages across three passage types. For \textbf{quantized models}, we report percentage point change relative to the unquantized model.}
  \label{tab:quant-dp-results}
\end{table}

\begin{table}[t]
\centering
\small
\resizebox{\linewidth}{!}{%
    \begin{tabular}{llccc}
    \toprule
    \textbf{Model} & \textbf{Group} & \textbf{English} & \textbf{Official Trans.} & \textbf{Unseen Trans.} \\
    \midrule
    \multirow{3}{*}{\textsc{LLaMA 3.1 8B}} 
    & Baseline      & 8.5\% & 3.1\% & 0.9\% \\
    & + w4a16       & \textbf{-0.7\%} & \textbf{-0.7\%} & \textbf{+0.0\%} \\
    & + w8a16       & \textbf{-0.2\%} & \textbf{-0.2\%} & \textbf{+0.0\%} \\
    \midrule
    \multirow{3}{*}{\textsc{LLaMA 3.1 70B}} 
    & Baseline      & 23.3\% & 9.0\% & 1.8\% \\
    & + w4a16       & \textbf{-1.2\%} & \textbf{-0.9\%} & \textbf{+0.1\%} \\
    & + w8a16       & \textbf{-5.8\%} & \textbf{-5.3\%} & \textbf{-1.3\%} \\
    \bottomrule
    \end{tabular}
}
\caption{\textbf{Name Cloze} accuracy for LLaMA 3.1 models (8B and 70B) grouped by language setting. For quantized models, we report percentage point change relative to the unquantized baseline.}
\label{tab:quant-nct-results}
\end{table}

\begin{table}[t]
\centering
\small
\begin{tabular}{llcccc}
\toprule
\textbf{Model} & \textbf{Condition} & \textbf{English} & \textbf{Translations} \\
\midrule
\multirow{3}{*}{\textsc{LLaMA 3.1 8B}} 
& Baseline      & 22.3\% & 20.1\% \\
& + w4a16       & \textbf{-0.3\%} & \textbf{-0.3\%} \\
& + w8a16       & \textbf{-0.1\%} & \textbf{-0.1\%} \\
\midrule
\multirow{3}{*}{\textsc{LLaMA 3.1 8B}} 
& No NE         & 22.3\% & 19.8\% \\
& + w4a16       & \textbf{-0.2\%} & \textbf{-0.3\%} \\
& + w8a16       & \textbf{+0.0\%} & \textbf{-0.1\%} \\
\midrule
\multirow{3}{*}{\textsc{LLaMA 3.1 70B}} 
& Baseline      & 25.4\% & 20.4\% \\
& + w4a16       & \textbf{-1.3\%} & \textbf{-0.3\%} \\
& + w8a16       & \textbf{-1.4\%} & \textbf{-0.6\%} \\
\midrule
\multirow{3}{*}{\textsc{LLaMA 3.1 70B}} 
& No NE         & 24.1\% & 20.7\% \\
& + w4a16       & \textbf{-0.5\%} & \textbf{-0.2\%} \\
& + w8a16       & \textbf{-0.8\%} & \textbf{-1.1\%} \\
\bottomrule
\end{tabular}
\caption{\textbf{Prefix Probe} accuracy (measured by ChrF++) for LLaMA 3.1 models (8B and 70B) on Standard and NE-removed (No NE) passages across English and Translation groups. Quantized model scores are reported as percentage point change relative to the full-precision baseline.}
\label{tab:quant-pp-results}
\end{table}

\begin{table*}[t]
  \centering
  \small
  \setlength{\tabcolsep}{4pt}
  \newcolumntype{L}[1]{>{\raggedright\arraybackslash}p{#1}}
  \newcolumntype{C}[1]{>{\centering\arraybackslash}p{#1}}
  \begin{tabularx}{\textwidth}{L{3.75cm} C{1.75cm} C{1.75cm} C{1.75cm} C{1.75cm} C{1.75cm} C{1.75cm}}
    \toprule
    \textbf{Index} & \textbf{No NC Official Translation} & \textbf{No NC Original English} & \textbf{No NC Unseen Translation} & \textbf{One NC Official Translation} & \textbf{One NC Original English} & \textbf{One NC Unseen Translation} \\\midrule
    v4\_c4train\_llama & 0 & 595 & 3 & 0 & 639 & 2 \\
    v4\_dclm-baseline\_llama & 3 & 1245 & 3 & 1 & 1266 & 3 \\
    v4\_dolma-v1\_6-sample\_llama & 0 & 36 & 0 & 0 & 36 & 0 \\
    v4\_dolma-v1\_7\_llama & 11 & 1226 & 3 & 16 & 1225 & 5 \\
    v4\_dolmasample\_olmo & 0 & 0 & 0 & 0 & 0 & 0 \\
    v4\_olmo-2-0325-32b-instruct\_llama & 6 & 1275 & 3 & 9 & 1304 & 3 \\
    v4\_olmo-2-1124-13b-instruct\_llama & 6 & 1275 & 3 & 9 & 1304 & 3 \\
    v4\_olmo-mix-1124\_llama & 3 & 1274 & 3 & 1 & 1300 & 3 \\
    v4\_olmoe-0125-1b-7b-instruct\_llama & 6 & 1275 & 3 & 9 & 1304 & 3 \\
    v4\_piletrain\_llama & 248 & 1307 & 2 & 249 & 1371 & 0 \\
    v4\_pileval\_gpt2 & 0 & 0 & 0 & 0 & 0 & 0 \\
    v4\_pileval\_llama & 0 & 73 & 0 & 0 & 63 & 0 \\
    v4\_rpj\_llama\_s4 & 247 & 1372 & 3 & 249 & 1425 & 2 \\
    \bottomrule
  \end{tabularx}
  \caption{\textbf{Infinigram Search} results by language on passages without a character name (Non NC) and with a character name (One NC). For unseen translation passages that were found, note that the chunks found within each passage are exclusively English leakage, not translated, unseen language text. We mark a passage as \emph{seen} (i.e., present in the training data) if it contains at least one matching span of $\leq$20 words; otherwise, we label it as \emph{unclear} and exclude it from the analysis. }
\label{tab:infinigram-non-ne}
\end{table*}

\section{Book-Level Accuracy Visualizations}
\label{sec:book-level-heatmaps}

\begin{figure*}[t]
    \centering
    \includegraphics[width=\textwidth]{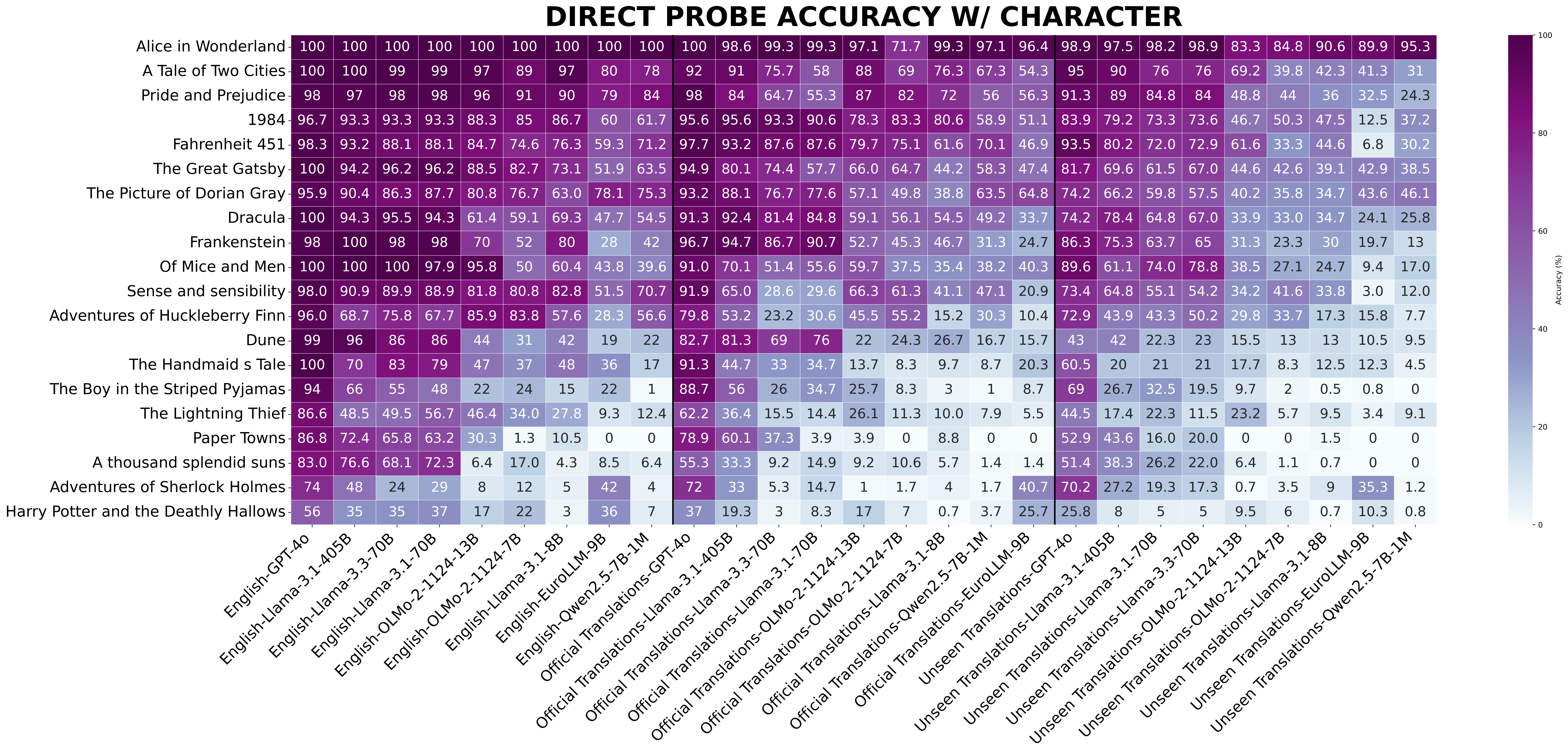}
    \caption{\textbf{Direct Probe} accuracy on unmasked passages containing a named entity of type Person. Rows correspond to individual book titles, sorted top-to-bottom by average model performance. Columns represent language/model combinations grouped into three regions: English (left), Official Translations (center), and Unseen Translations (right). Accuracy is reported as a percentage.}

    \label{fig:heatmap-unmasked}
\end{figure*}

\begin{figure*}[t]
    \centering
    \includegraphics[width=\textwidth]{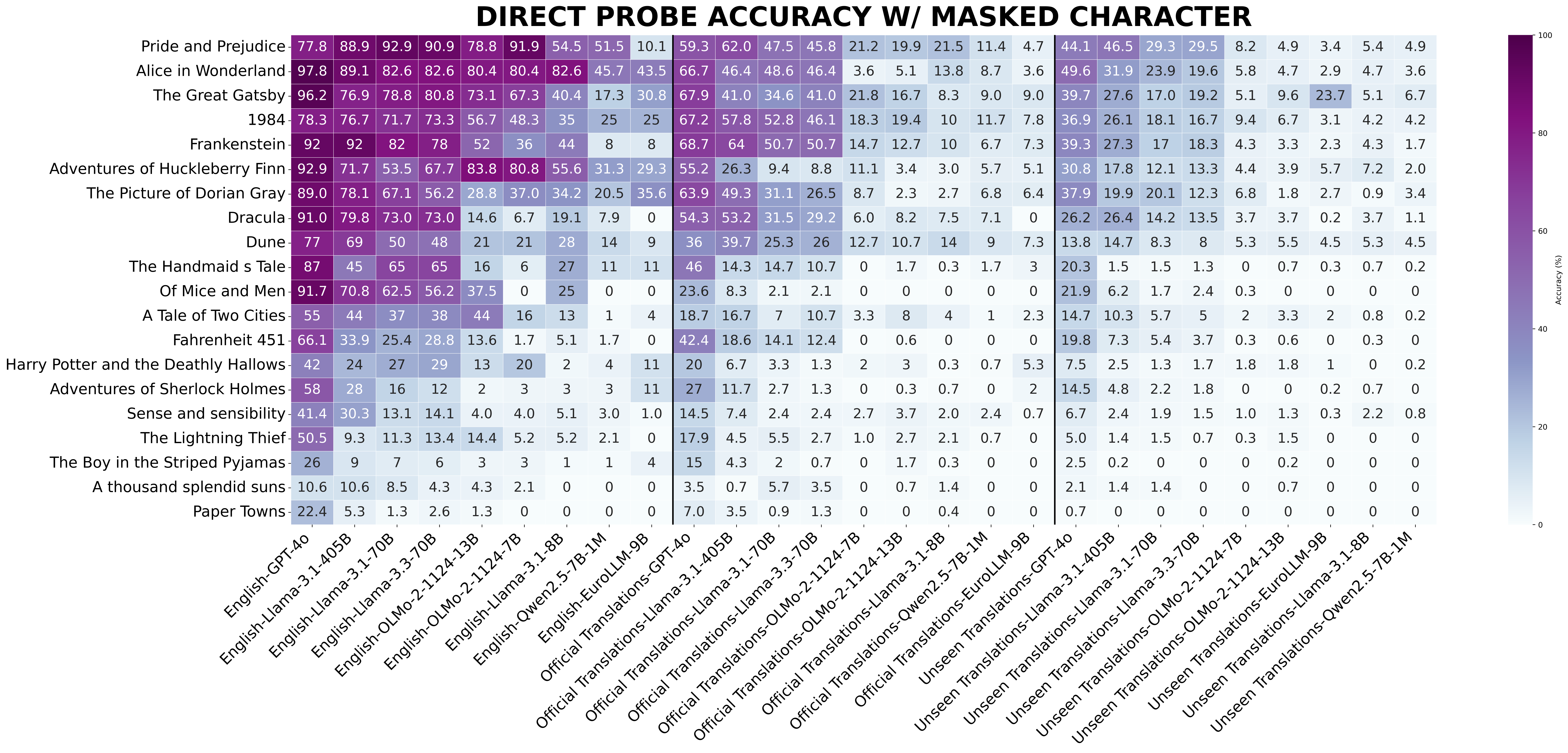}
    \caption{\textbf{Direct Probe} accuracy on masked passages where the single named character has been replaced with [MASK]. Books are sorted by overall average accuracy (top-to-bottom), and models are grouped by language setting: English, Official Translations, and Unseen Translations. Accuracy values are shown as percentages.}
    \label{fig:heatmap-masked}
\end{figure*}

\begin{figure*}[t]
    \centering
    \includegraphics[width=\textwidth]{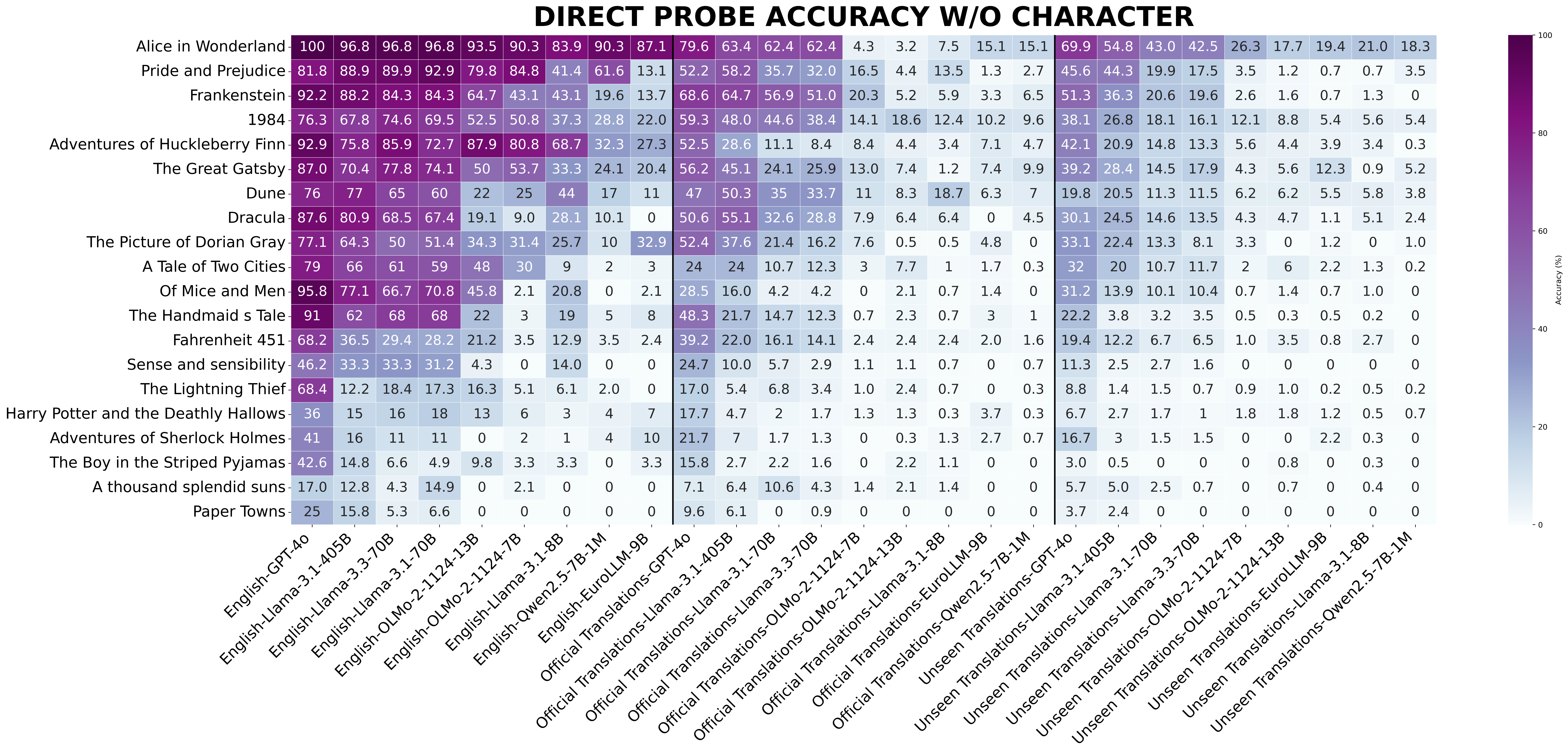}
    \caption{\textbf{Direct Probe} accuracy on passages without any named entity of type Person. Rows indicate books (sorted by average performance), and columns are grouped by language category: English, official translations, and unseen translations (newly produced machine translations). Values represent accuracy percentages.}
    \label{fig:heatmap-non-ne}
\end{figure*}

\begin{figure*}[t]
    \centering
    \includegraphics[width=\textwidth]{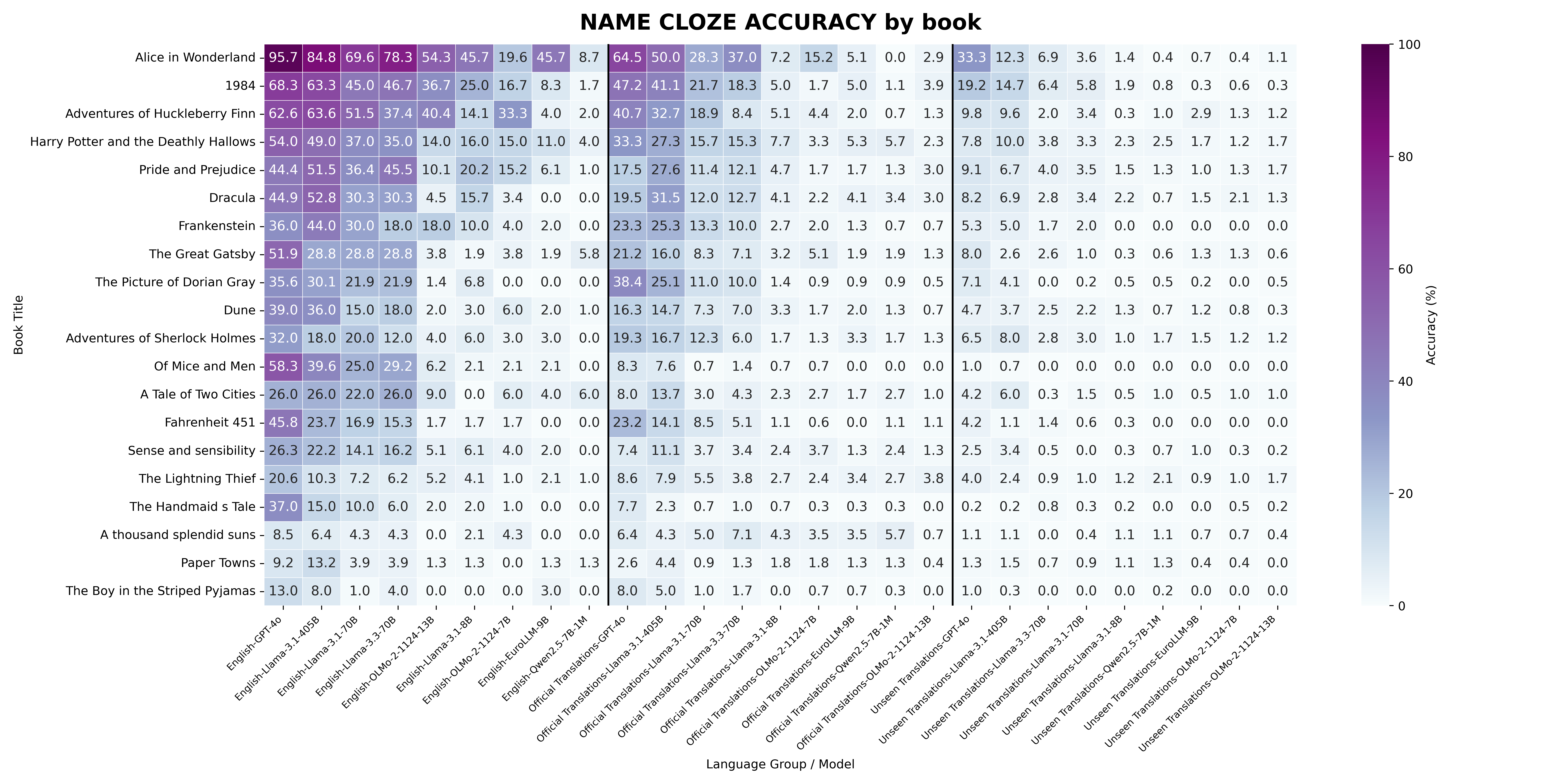}
    \caption{\textbf{Name Cloze} accuracy by book. Each row represents a title (sorted by average performance), and columns show performance across models grouped by language: English (left), official translations (center), and unseen translations (right). Accuracy is computed as the percentage of correct predictions.}
    \label{fig:heatmap-nct}
\end{figure*}

\begin{figure*}[t]
    \centering
    \includegraphics[width=\textwidth]{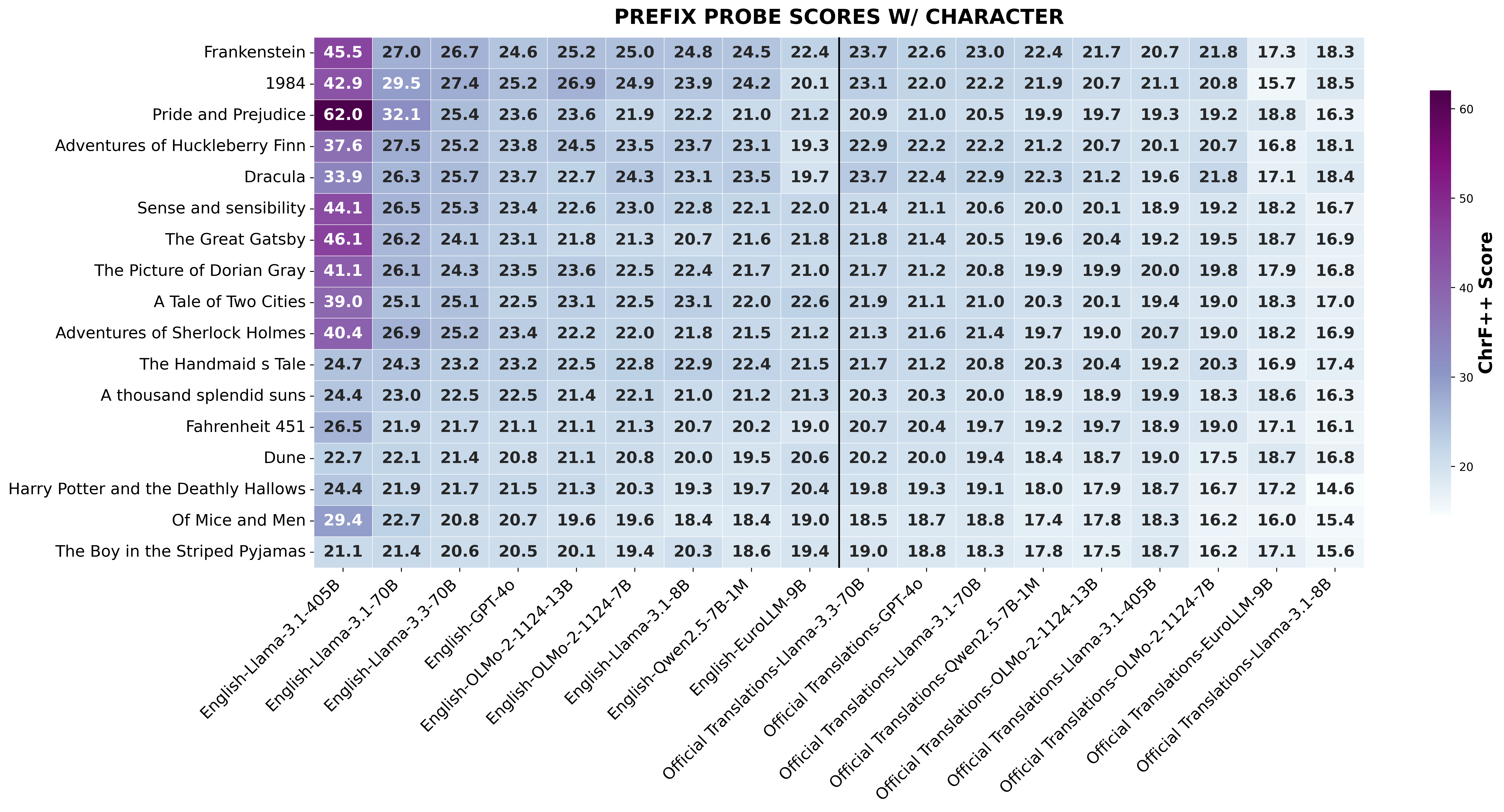}
    \caption{\textbf{Prefix Probe} score on unmasked passages containing a named entity of type Person. Each row represents a title (sorted by average performance), and columns show performance across models grouped by language: English (left), and official translations (right). ChrF++ scores are computed as character-level overlap between model-generated text continuations and ground truth passages.}
    \label{fig:heatmap-pp-w-char}
\end{figure*}

\begin{figure*}[t]
    \centering
    \includegraphics[width=\textwidth]{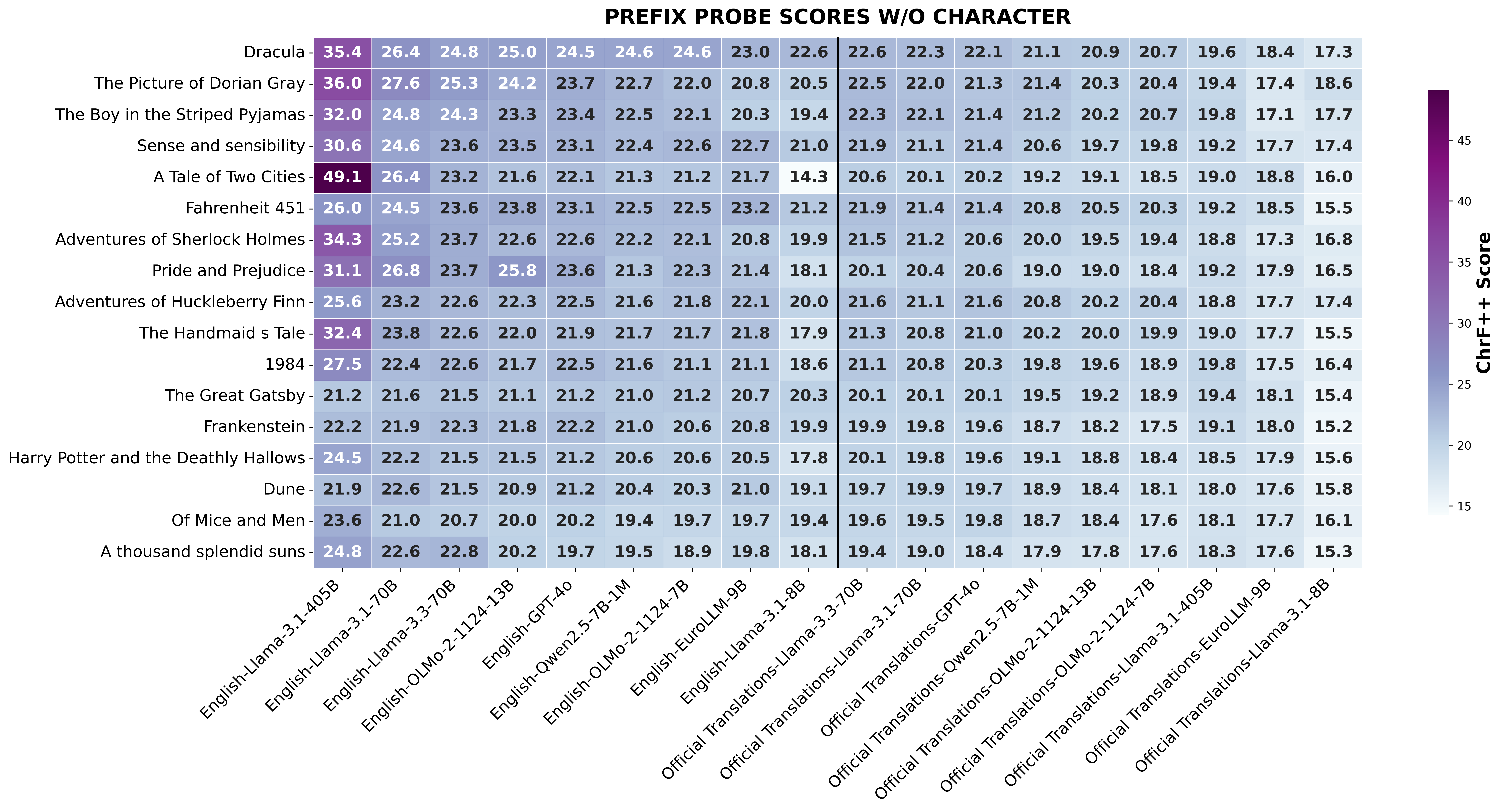}
    \caption{\textbf{Prefix Probe} score on unmasked passages without any named entity of type Person. Each row represents a title (sorted by average performance), and columns show performance across models grouped by language: English (left), and Official Translations (right). ChrF++ scores are computed as character-level overlap between model-generated text continuations and ground truth passages.}
    \label{fig:heatmap-pp-wo-char}
\end{figure*}

To better understand how memorization patterns vary across individual titles, we visualize model performance at the book level for each probing task and setting. Figures \ref{fig:heatmap-unmasked}, \ref{fig:heatmap-masked}, \ref{fig:heatmap-non-ne}, \ref{fig:heatmap-nct}, \ref{fig:heatmap-pp-w-char}, and \ref{fig:heatmap-pp-wo-char} display accuracy heatmaps for Direct Probing, Name Cloze, and Prefix Probe, broken down by book title, language group, and model.

\begin{itemize}
    \item \textbf{\autoref{fig:heatmap-unmasked}} shows Direct Probe accuracy on standard passages containing character name.
    \item \textbf{\autoref{fig:heatmap-masked}} reports Direct Probe accuracy when the named entity is masked from the passage.
    \item \textbf{\autoref{fig:heatmap-non-ne}} displays Direct Probe accuracy on passages without character names.
    \item \textbf{\autoref{fig:heatmap-nct}} visualizes Name Cloze accuracy, where the model must recover the correct character name from a passage with masked passage.
    \item \textbf{\autoref{fig:heatmap-pp-w-char}} shows Prefix Probe accuracy on standard passages containing character name. Accuracy is reported as mean chrF++.
    \item \textbf{\autoref{fig:heatmap-pp-wo-char}} reports Prefix Probe accuracy on passages without character names. Accuracy is reported as mean chrF++.
\end{itemize}

These visualizations reveal substantial variation in model behavior across books. High memorization rates on well-known titles like \textit{Alice in Wonderland} or \textit{Of Mice and Men} contrast sharply with near-zero accuracy on less culturally prominent works or in unseen translation settings. They also highlight the sensitivity of LLM recall to entity presence and surface form, which is less apparent in aggregate-level analyses.

\begin{figure*}[t]
    \centering
    \includegraphics[width=\textwidth]{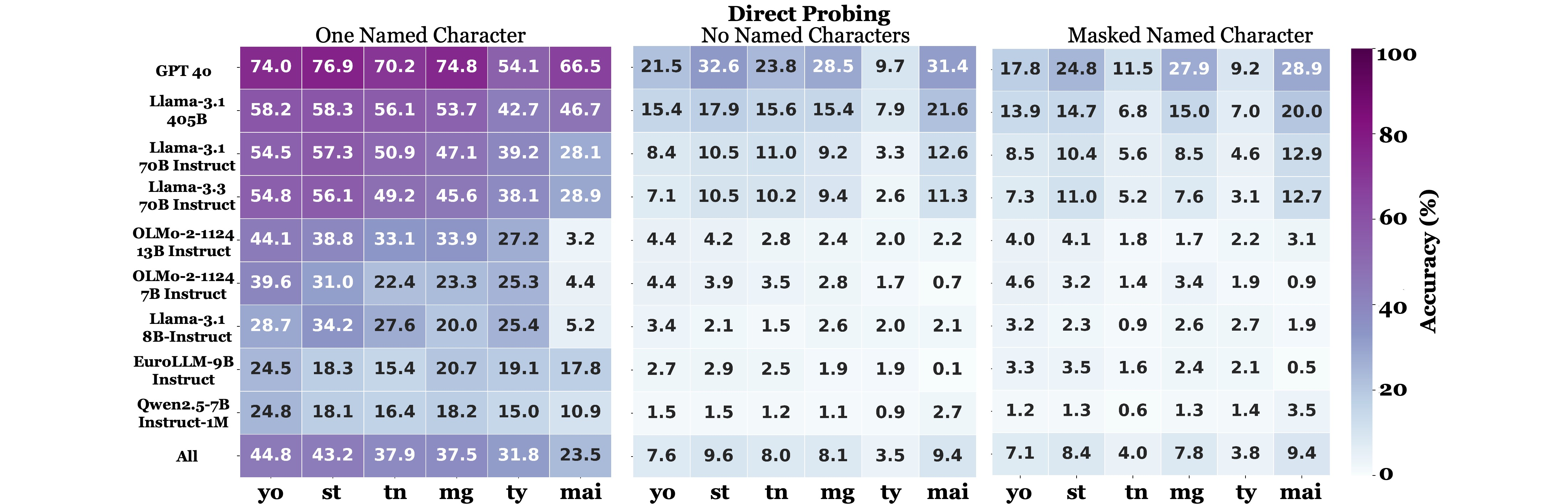}
    \caption{\textbf{Direct Probe} accuracy across different settings on newly produced machine translations: (1) passages with a character name, (2) passages without a character name, and passages where that name was masked. Accuracy values are shown as percentages.}
    \label{fig:heatmap-dp-clm}
\end{figure*}

\begin{table*}[t]
\centering
\footnotesize
\rowcolors{2}{gray!15}{white}
\begin{tabular}{lcccccc}
\toprule
\textbf{Language} & \multicolumn{2}{c}{\textbf{Masked Entity}} & \multicolumn{2}{c}{\textbf{No Character}} & \multicolumn{2}{c}{\textbf{Unmasked Entity}} \\
& Author Correct & Suspicious & Author Correct & Suspicious & Author Correct & Suspicious \\
\toprule
\textbf{English}  & 0.23 & 0.05 & 0.21 & 0.07 & 0.36 & 0.10 \\
\textbf{Spanish}  & 0.10 & 0.07 & 0.08 & 0.10 & 0.29 & 0.12 \\
\textbf{Turkish}  & 0.09 & 0.08 & 0.07 & 0.13 & 0.35 & 0.15 \\
\textbf{Vietnamese}  & 0.08 & 0.11 & 0.06 & 0.19 & 0.31 & 0.23 \\
\textbf{Maithili} & 0.06 & 0.51 & 0.05 & 0.95 & 0.12 & 0.60 \\
\textbf{Sesotho}  & 0.04 & 0.16 & 0.04 & 0.34 & 0.21 & 0.32 \\
\textbf{Yoruba}  & 0.04 & 0.19 & 0.04 & 0.40 & 0.24 & 0.40 \\
\textbf{Malagasy} & 0.04 & 0.62 & 0.04 & 1.12 & 0.20 & 0.88 \\
\textbf{Tswana}  & 0.02 & 0.33 & 0.04 & 0.59 & 0.18 & 0.56 \\
\textbf{Tahitian}  & 0.01 & 0.45 & 0.02 & 0.84 & 0.15 & 0.62 \\
\bottomrule
\end{tabular}
\caption{Percentage of only author being correct and response being an erroneous text (i.e "unknown"," ", "none", "book name") with respect to total incorrect answers in that language.}
\end{table*}

\begin{table*}[t]
\centering
\footnotesize
\rowcolors{2}{gray!15}{white}
\begin{tabular}{lrrr}
\toprule
\textbf{Model} & \textbf{Masked character} & \textbf{No character} & \textbf{W/ character} \\
\midrule
EuroLLM-9B-Instruct           & 3905 & 4720 & 4691 \\
Meta-Llama-3.1-8B-Instruct    & 2279 & 2960 & 1274 \\
Llama-3.3-70B-Instruct        & 1321 & 3663 & 1006 \\
Qwen2.5-7B-Instruct-1M        &  289 &  790 &  494 \\
OLMo-2-1124-13B-Instruct      &  181 &  738 &  209 \\
Llama-3.1-405B                &   67 &   38 &   14 \\
Llama-3.1-70B-Instruct        &   32 & 1188 &   57 \\
Qwen-2.5-Omni-7b              &   28 &   32 &   12 \\
GPT-4o                        &   25 &   16 &   24 \\
OLMo-2-1124-7B-Instruct       &   16 &  280 &  107 \\
\bottomrule
\end{tabular}
\caption{\textbf{Direct probing errors:} Number of responses where the model abstained or did not complete the task, returning either an empty string or one of the following: "unknown", "none", "book name", "author name".}
\end{table*}

\begin{table*}[t]
\centering
\footnotesize
\renewcommand{\arraystretch}{1.3}
\rowcolors{2}{gray!15}{white}
\resizebox{\textwidth}{!}{%
\begin{tabular}{clrlrlrl}
\midrule
\rowcolor{violet!30}
\textbf{Lang} & \textbf{Title \& Author (masked chatacter)} & \textbf{Count} & \textbf{Title \& Author (w/o character)} & \textbf{Count} & \textbf{Title \& Author (w/ character)} & \textbf{Count} \\
\toprule
en & "Pride And Prejudice", "Jane Austen" & 535 & "Pride And Prejudice", "Jane Austen" & 436 & "Alice's Adventures In Wonderland", "Lewis Carroll" & 277 \\
en & "The Catcher In The Rye", "J.D. Salinger" & 292 & "The Catcher In The Rye", "J.D. Salinger" & 258 & "The Hound Of The Baskervilles", "Arthur Conan Doyle" & 178 \\
en & "The Adventures Of Tom Sawyer", "Mark Twain" & 272 & "The Hound Of The Baskervilles", "Arthur Conan Doyle" & 215 & "The Adventures Of Tom Sawyer", "Mark Twain" & 148 \\
\midrule
es & "Don Quixote", "Miguel De Cervantes" & 726 & "El Señor De Los Anillos", "J.R.R. Tolkien" & 847 & "El Señor De Los Anillos", "J.R.R. Tolkien" & 431 \\
es & "El Señor De Los Anillos", "J.R.R. Tolkien" & 599 & "Don Quixote", "Miguel De Cervantes" & 473 & "Harry Potter Y El Prisionero De Azkaban", "J.K. Rowling" & 164 \\
es & "Cien Años De Soledad", "Gabriel García Márquez" & 313 & "La Sombra Del Viento", "Carlos Ruiz Zafón" & 310 & "The Hound Of The Baskervilles", "Arthur Conan Doyle" & 147 \\
\midrule
vi & "The Secret Garden", "Frances Hodgson Burnett" & 596 & "The Secret Garden", "Frances Hodgson Burnett" & 473 & "The Scarlet Letter", "Nathaniel Hawthorne" & 288 \\
vi & "The Kite Runner", "Khaled Hosseini" & 529 & "The Kite Runner", "Khaled Hosseini" & 392 & "The Catcher In The Rye", "J.D. Salinger" & 217 \\
vi & "The Scarlet Letter", "Nathaniel Hawthorne" & 466 & "The Catcher In The Rye", "J.D. Salinger" & 343 & "The Hound Of The Baskervilles", "Arthur Conan Doyle" & 205 \\
\midrule
tr & "The Count Of Monte Cristo", "Alexandre Dumas" & 610 & "The Count Of Monte Cristo", "Alexandre Dumas" & 450 & "Harry Potter", "J.K. Rowling" & 200 \\
tr & "Moby Dick", "Herman Melville" & 562 & "Crime And Punishment", "Fyodor Dostoevsky" & 437 & "Alice's Adventures In Wonderland", "Lewis Carroll" & 179 \\
tr & "Crime And Punishment", "Fyodor Dostoevsky" & 319 & "Moby Dick", "Herman Melville" & 302 & "Aşk Ve Gurur", "Jane Austen" & 179 \\
\midrule
mai & "The Scarlet Letter", "Nathaniel Hawthorne" & 783 & "The Scarlet Letter", "Nathaniel Hawthorne" & 577 & "The Scarlet Letter", "Nathaniel Hawthorne" & 1034 \\
mai & "To Kill A Mockingbird", "Harper Lee" & 699 & "To Kill A Mockingbird", "Harper Lee" & 422 & "Pride And Prejudice", "Jane Austen" & 422 \\
mai & "Pride And Prejudice", "Jane Austen" & 675 & "The Jungle Book", "Rudyard Kipling" & 370 & "The Jungle Book", "Rudyard Kipling" & 383 \\
\midrule
mg & "The Scarlet Letter", "Nathaniel Hawthorne" & 609 & "The Count Of Monte Cristo", "Alexandre Dumas" & 558 & "The Scarlet Letter", "Nathaniel Hawthorne" & 285 \\
mg & "To Kill A Mockingbird", "Harper Lee" & 570 & "To Kill A Mockingbird", "Harper Lee" & 504 & "Les Misérables", "Victor Hugo" & 228 \\
mg & "The Count Of Monte Cristo", "Alexandre Dumas" & 528 & "The Scarlet Letter", "Nathaniel Hawthorne" & 421 & "Alice's Adventures In Wonderland", "Lewis Carroll" & 217 \\
\midrule
st & "To Kill A Mockingbird", "Harper Lee" & 1199 & "To Kill A Mockingbird", "Harper Lee" & 691 & "Alice's Adventures In Wonderland", "Lewis Carroll" & 262 \\
st & "The Lord Of The Rings", "J.R.R. Tolkien" & 676 & "The Lord Of The Rings", "J.R.R. Tolkien" & 646 & "Harry Potter And The Philosopher's Stone", "J.K. Rowling" & 256 \\
st & "Moo", "Sol Plaatje" & 415 & "Moo", "Sol Plaatje" & 476 & "To Kill A Mockingbird", "Harper Lee" & 212 \\
\midrule
tn & "To Kill A Mockingbird", "Harper Lee" & 1656 & "The No. 1 Ladies' Detective Agency", "Alexander McCall Smith" & 955 & "The No. 1 Ladies' Detective Agency", "Alexander McCall Smith" & 341 \\
tn & "The No. 1 Ladies' Detective Agency", "Alexander McCall Smith" & 876 & "To Kill A Mockingbird", "Harper Lee" & 795 & "To Kill A Mockingbird", "Harper Lee" & 290 \\
tn & "Moo", "Sol Plaatje" & 644 & "Mafingwane", "Thomas Mofolo" & 245 & "Alice's Adventures In Wonderland", "Lewis Carroll" & 229 \\
\midrule
ty & "Moby-Dick", "Herman Melville" & 1174 & "Moby-Dick", "Herman Melville" & 834 & "The Scarlet Letter", "Nathaniel Hawthorne" & 536 \\
ty & "The Lord Of The Rings", "J.R.R. Tolkien" & 575 & "Leaves Of Grass", "Walt Whitman" & 478 & "Moby-Dick", "Herman Melville" & 346 \\
ty & "To Kill A Mockingbird", "Harper Lee" & 457 & "The Pearl", "John Steinbeck" & 295 & "The Lord Of The Rings", "J.R.R. Tolkien" & 282 \\
\midrule
yo & "Things Fall Apart", "Chinua Achebe" & 2969 & "Things Fall Apart", "Chinua Achebe" & 3226 & "Things Fall Apart", "Chinua Achebe" & 762 \\
yo & "To Kill A Mockingbird", "Harper Lee" & 533 & "The Palm-Wine Drinkard", "Amos Tutuola" & 462 & "Alice's Adventures In Wonderland", "Lewis Carroll" & 254 \\
yo & "Things Fall Apart", "Chinua Achebe" & 370 & "title": "the lion and the jewel","author": "wole soyinka" & 326 & "Harry Potter And The Philosopher's Stone", "J.K. Rowling" & 218 \\
\bottomrule
\end{tabular}
}
\caption{\textbf{Direct probing errors:} The three most frequently returned incorrect titles and authors, with their respective counts shown per language and across the three evaluation settings.}
\end{table*}

\begin{table*}[t]
\centering
\footnotesize
\renewcommand{\arraystretch}{1.3}
\rowcolors{2}{gray!15}{white}
\resizebox{\textwidth}{!}{
  \begin{tabular}{lrrrrr}
    \toprule
    \rowcolor{violet!30}
    \textbf{Language} & \textbf{[MASK]} & \textbf{Unknown/\texttt{name}} & \textbf{Pronoun} & \textbf{Honorific} & \textbf{Another Name} \\
    \midrule
    en   & 0.015 & 0.008 & 0.077 & 0.122 & 0.778 \\
    es   & 0.027 & 0.001 & 0.057 & 0.092 & 0.823 \\
    vi   & 0.002 & 0.002 & 0.039 & 0.025 & 0.932 \\
    tr   & 0.009 & 0.001 & 0.015 & 0.037 & 0.938 \\
    yo   & 0.001 & 0     & 0.004 & 0.017 & 0.978 \\
    mg   & 0.001 & 0     & 0.002 & 0.019 & 0.977 \\
    mai  & 0.003 & 0.001 & 0.004 & 0.018 & 0.974 \\
    tn   & 0.012 & 0     & 0.009 & 0.011 & 0.968 \\
    st   & 0.001 & 0     & 0.003 & 0.021 & 0.976 \\
    ty   & 0     & 0.001 & 0.007 & 0.006 & 0.987 \\
    \rowcolor{gray!10}
    \textbf{Total} & 0.007 & 0.001 & 0.021 & 0.036 & 0.935 \\
    \bottomrule
  \end{tabular}
}
\caption{\textbf{Name Cloze:} Breakdown of incorrect character predictions per language. Columns indicate the count of [MASK] returns, unknown/name tokens, pronouns, honorifics, and alternative names. Top 4 most frequently returned names per language are also listed with counts.}
\label{tab:name_breakdown}
\end{table*}

\begin{table*}[ht]
\centering
\scriptsize
\renewcommand{\arraystretch}{1.2}
\begin{tabularx}{\textwidth}{|p{3.5cm}|X|}
\midrule
\rowcolor{gray!15}
\textbf{Error Type} & \textbf{Description} \\
\midrule
\rowcolor{yellow!20}
\textsc{Wrong Title and Author} &
\textbf{Definition:} Model returns an unrelated, but often famous, title-author pair. \\
& \textbf{Example:} \texttt{"title": "Altered Carbon", "author": "Richard K. Morgan"} \\
& \textbf{Correct answer} \texttt{Dune}\\
& \textbf{Model} \texttt{Olmo2-1124-13B-Instruct}\\
& \textbf{Task:} Direct Probe \\

\midrule
\rowcolor{yellow!20}
\textsc{Correct Author, Wrong Title} &
\textbf{Definition:} Author is correctly identified, but the title is incorrect.\footnote{The opposite of this is not seen.} \\

& \textbf{Example:} \texttt{ "title": "Dune Messiah", "author": "Frank Herbert"}\\
& \textbf{Correct:} \texttt{ "title": "Dune","author":"Frank Herbert"}\\
& \textbf{Model} \texttt{Olmo2-1124-13B-Instruct}\\
& \textbf{Task:} Direct Probe \\


\midrule
\rowcolor{blue!10}
\textsc{Refusal or Abstention} &
\textbf{Definition:} Model fails to make a guess, returning ``Unknown'' or similar. \\
& \textbf{Example:} \texttt{"title": "Book name: Unknown", "author": "Unknown author"} \\
& \textbf{Correct:} \texttt{title: Dune author : Frank Herbert}\\
& \textbf{Model}: \texttt{Llama-3.1-8B-Instruct}\\
& \textbf{Task:}\texttt{ Direct Probe} \\

\midrule
\rowcolor{yellow!20}
\textsc{Wording or Stylistic Errors} &
\textbf{Definition:} Title is misworded, reformatted, or awkwardly phrased. \\
& \textbf{Example:} \texttt{"""title"": ""Nineteen Eighty-Four"", ""author"": ""George Orwell"""}\\
& \textbf{Correct Answer:} \texttt{title : 1984, author: George Orwell} \\
& \textbf{Model} \texttt{Gpt-4o-audio-preview}\\
& \textbf{Task:} Direct Probe \\

\midrule
\rowcolor{yellow!20}
\textsc{Incorrect Entity from Same Book} &
\textbf{Definition:} Returns a different character from the same book.\\
& \textbf{Example:} \texttt{Charles}\\
& \textbf{Correct Answer}: \texttt{Mr.Lorry} \\
& \textbf{Model:} \texttt{Llama3.1-405b}\\
& \textbf{Task:} \texttt{Name Cloze} \\

\midrule
\rowcolor{yellow!20}
\textsc{Culturally Popular but Incorrect Name} &
\textbf{Definition:} Model selects an incorrect name which is specific to the culture of the passage language. \\
& \textbf{Example:} \texttt{``Ataturk''} \\
& \textbf{Correct answer:} \texttt{Winston} \\
& \textbf{Model} : \texttt{LLama-3.3-70B}\\
& \textbf{Task:} \texttt{Name Cloze} \\

\midrule
\rowcolor{blue!10}
\textsc{Multi-Guess Output} &
\textbf{Definition:} Model provides multiple candidates or alternative guesses. \\
& \textbf{Example:} Model response: \\
& \texttt{Based on the context of the passage, I'm going to take a guess that the proper name that fills the [MASK] token is: Fahrenheit.} \\
& \texttt{However, this seems unlikely, as "Fahrenheit" is a title of a book, not a character's name. A more plausible guess would be a character from a dystopian novel, such as "Fahrenheit 451".} \\
& \texttt{Mildred} \\
& \textbf{Correct Answer:} \texttt{Hermione} \\
& \textbf{Model:} Llama3.1-405b  \\
& \textbf{Task:} Name Cloze \\
\midrule

\midrule
\rowcolor{green!10}
\textsc{Broken or Corrupted Output} &
\textbf{Definition:} Model outputs unreadable, fragmented, or nonsensical tokens. \\
& \textbf{Example:} \texttt{"title": ".k%
\begin{CJK}{UTF8}{min}ハウス\end{CJK} absorbing riches%
\begin{CJK}{UTF8}{min}て舟へ\end{CJK}.", 
"author": "%
\begin{CJK}{UTF8}{min}魚つり人才育てる\end{CJK}(Balls to Become a Fishing Pro %
\begin{CJK}{UTF8}{min}魚つり人才を作り出す\end{CJK}!)"}\\
& \textbf{Correct Answer:} Marianne\\ 
& \textbf{Model:} Qwen-2.5-Omni-7b \\
& \textbf{Task:} Both \\

\midrule
\rowcolor{blue!10}
\textsc{Honorific or Pronoun returned}&
\textbf{Definition:} Model outputs a Honorific or Pronoun instead of entity \\
& \textbf{Example:} \texttt{Mr.} \\&
\textbf{Correct Answer:} Mr. Darcy \\
&\textbf{Model:} Llama-3.1-8B-Instruct \\
&\textbf{Task:} Both\\
\midrule
\end{tabularx}
\caption{Defined error types with descriptions, examples, and applicable tasks}
\label{tab:custom_error_types}
\end{table*}